\documentclass[letterpaper, 10 pt, conference]{ieeeconf}  

\IEEEoverridecommandlockouts                              

\usepackage{amsmath} 
\usepackage{amssymb}  

\usepackage{graphicx}
\usepackage{subfigure}
\newcommand{\ourmethod}{\text{DMPO}}

\newtheorem{mythm}{Theorem}

\newtheorem{myremark}{Remark}

\usepackage[noend]{algorithmic}
\usepackage[ruled,vlined,linesnumbered]{algorithm2e}

\usepackage{diagbox}
\usepackage{url}
\usepackage{balance}
\usepackage{color}
\newcommand{\yali}[1]{{ \color{magenta}[Yali: #1]}}

\usepackage{changes}

\title{\LARGE \bf
Scalable Model-based Policy Optimization for \\ Decentralized Networked Systems
}

\author{Albert Author$^{1}$ and Bernard D. Researcher$^{2}$
\thanks{*This work was not supported by any organization}
\thanks{$^{1}$Albert Author is with Faculty of Electrical Engineering, Mathematics and Computer Science,
        University of Twente, 7500 AE Enschede, The Netherlands
        {\tt\small albert.author@papercept.net}}%
\thanks{$^{2}$Bernard D. Researcheris with the Department of Electrical Engineering, Wright State University,
        Dayton, OH 45435, USA
        {\tt\small b.d.researcher@ieee.org}}%
}

\author{Yali Du$^{1*}$,  Chengdong Ma$^{2*}$, Yuchen Liu$^{5}$, Runji Lin$^{3}$, Hao Dong$^{5}$, Jun Wang$^{4}$, Yaodong Yang$^{5,\dag}$
\thanks{*The first two authors contributed equally. $\dag$Correponding author.}
\thanks{$^{1}$ Yali Du is with King’s College London, 30 Aldwych, London, UK
        {\tt\small yali.du@kcl.ac.uk}}%
\thanks{$^{2}$ Chengdong Ma is with Xiamen University, 422 Siming Road, Xiamen, China
        {\tt\small machengdong@stu.xmu.edu.cn}. Work done as a research intern at Peking University}%
\thanks{$^{3}$ Runji Lin is with Chinese Academy of Sciences,  95 Zhongguancun East Road, Beijing, China
        {\tt\small linrunji2021@ia.ac.cn}}%
\thanks{$^{4}$ Jun Wang is with University College London, 66-72 Gower street, London, UK
        {\tt\small jun.wang@cs.ucl.ac.uk}}%
\thanks{$^{5}$ Yuchen Liu is with Peking University, 5 Yiheyuan Road, Haidian District, Beijing, China
        {\tt\small liuyuchen18@pku.edu.cn}}%
\thanks{$^{5}$ Hao Dong is with CFCS, School of CS, Peking University
        {\tt\small hao.dong@pku.edu.cn}}%
\thanks{$^{5}$ Yaodong Yang is with Institute for AI, Peking University \& BIGAI
        {\tt\small yaodong.yang@pku.edu.cn}}%
}

\begin{document}

\maketitle
\thispagestyle{empty}
\pagestyle{empty}

\begin{abstract}

Reinforcement learning algorithms  require a large amount of samples; this often limits their real-world applications on even  simple tasks. 
Such a challenge is more outstanding in multi-agent tasks, as each step of operation is more costly, requiring communications or shifting or resources.
This work aims to improve data efficiency of multi-agent control  by model-based learning. 
We consider networked systems where agents are cooperative and communicate only locally with their neighbors, and propose the  decentralized model-based policy optimization framework ($\ourmethod$).
In our method, each agent learns a dynamic model to predict future states and broadcast their predictions by communication, and then the policies are trained under the model rollouts. 
To alleviate the bias of model-generated data, we restrain the model usage for generating myopic rollouts, thus reducing the compounding error of model generation.
To pertain the independence of policy update, we introduce extended value function and theoretically prove that the resulting policy gradient is a close approximation to true policy gradients.
We evaluate our algorithm on several benchmarks for intelligent transportation systems, which are connected autonomous  vehicle  control  tasks (Flow and CACC) and adaptive  traffic  signal  control  (ATSC).
Empirical results show that our method achieves superior data efficiency and matches the  performance of model-free methods using true models.

The source code of our algorithm and baselines can be found at \url{https://github.com/PKU-MARL/Model-Based-MARL}.

\end{abstract}

\section{Introduction}

Many real world problems, such as autonomous driving, wireless communications, multi-player games can be modeled as multi-agent reinforcement learning (MARL) problems, where multiple autonomous agents coexist in a common environment, aiming to maximize its individual or team reward in the long term by interacting with the environment and other agents. Unlike single-agent tasks, multi-agent tasks are more challenging, due to partial observations and unstable environments when agents update their policies simultaneously. Therefore, there are hardly any one-fits-all solutions for MARL problems. For example, in networked systems control (NSC) \cite{chu2020multiagent}, agents are connected via a stationary network and perform decentralized control based on its local observations and messages from connected neighbors. Examples include connected vehicle control \cite{jin2014dynamics}, traffic signal control \cite{chu2020multiagent}, etc.

\begin{figure}[t!]
	\centering
	 	\includegraphics[width=0.65\columnwidth]{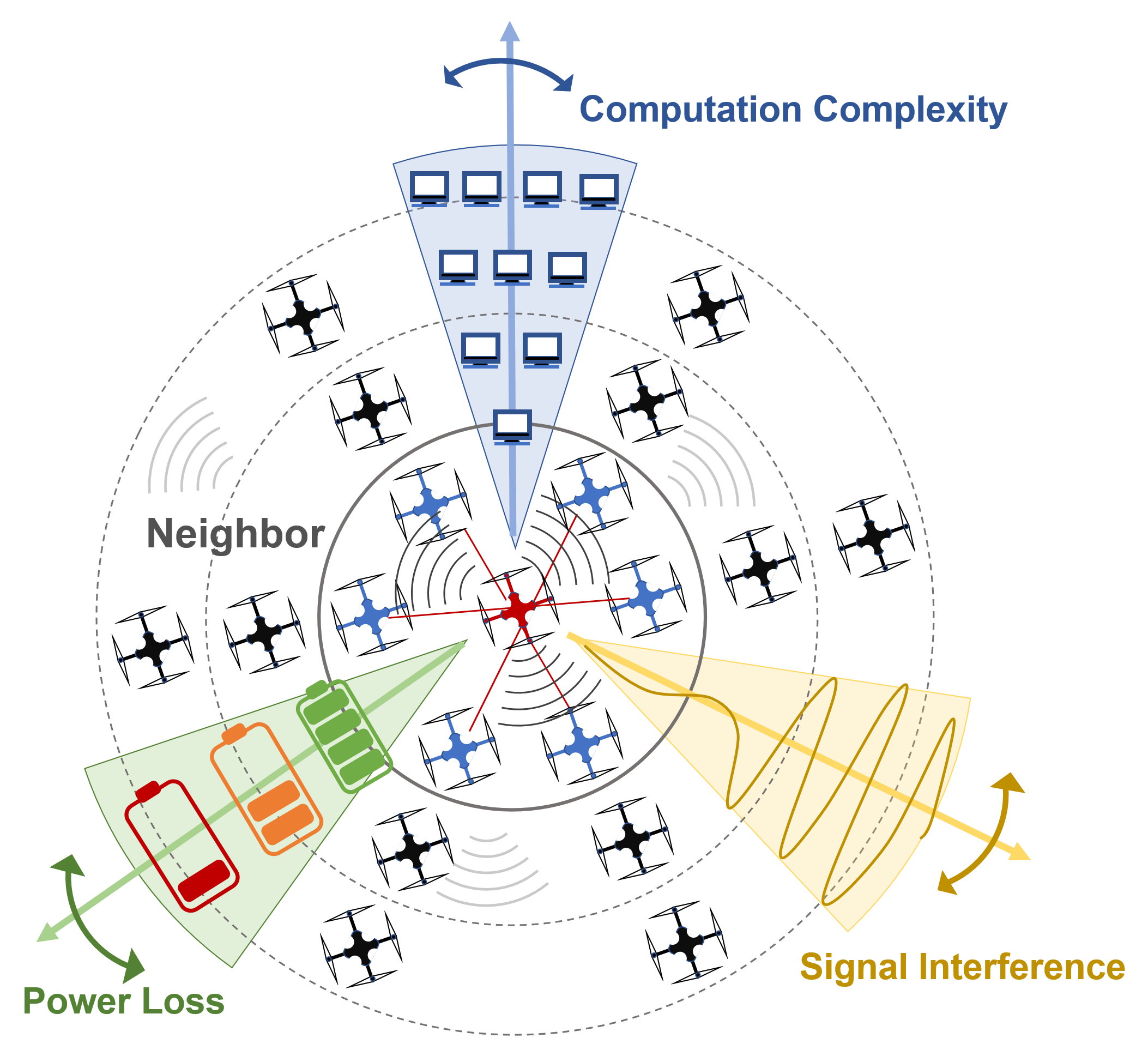}
    \caption{ Communication relationship of UAVs formation and disadvantages of massive communication operations.
    }
\label{fig:motivation}
\vspace{-0.2 in}
\end{figure}

{
Despite the emergence of many MARL algorithms \cite{chu2020multiagent,qu2020scalable,qu2020scalable2,yu2021surprising}, the data efficiency problem is often underestimated \cite{vinyals2019alphastar}.
On the one hand, in MARL tasks, agents are often connected via communication or coordination for cooperative control, making the decision making more complicated than single agent RL. 
On the other hand, many real applications demand high sample efficiency, preventing RL being applied. For example,  in networked storage  operation, one step of operations can be costly with loads shifted across a network and possible power loss \cite{qin2015distributed}. Figure \ref{fig:motivation} shows that in multiple unmanned aerial vehicle (UAV) formation, when executing military tasks, each step of communication operations will cause power loss and reduce the endurance of  UAVs. Frequent and massive communication operations  also increase the probability of  receiving signal interference for UAVs.
In contrast, model-based RL exploits an estimated dynamic model, is empirically more data-efficient than model-free approaches \cite{deisenroth2011pilco,deisenroth2013survey}. 
{While the success of model-based RL (MB-RL) has been witnessed in many single agent RL tasks \cite{Schrittwieser_2020,luo2019algorithmic,janner2019trust,morgan2021model},} the understanding of its MARL counterpart is still limited. 
Existing MB-MARL algorithms either limit their field of research on specific scenario, e.g. two-player zero-sum Markov game \cite{zhang2020model},  or tabular RL case \cite{bargiacchi2020model}. MB-MARL for multi-agent MDPs is still an open problem to be solved \cite{zhang2021multi}, 
{with profound challenges such as scalability issues caused by large state-action space and incomplete information of other agents’ state or actions \cite{wang2020model}.}}


In this paper, we restrict our attention to networked system control, where agents are able to communicate with others for the objective of cooperative control, 
{and propose the first decentralized model-based algorithm for networked systems, coined as  \textbf{D}ecentralized \textbf{M}odel-based \textbf{P}olicy \textbf{O}ptimization ($\ourmethod$), It is worth noting that the networked systems here are in a broad sense. Actually, $\ourmethod$ can be widely used to solve the control and decision-making problems of general multi-agent systems. Similarly, “decentralized” is in a broad sense for any multi-agent system in general. Our purpose is to improve the performance of the overall system when the information acquisition of a single agent is very limited.}
In $\ourmethod$, we use localized models to predict future states, and use communication to broadcast their predictions. To alleviate the issue of compounding model error, we adopt a branching strategy \cite{sutton1990integrated,janner2019trust} by replacing few long-horizon rollouts with many short-horizon rollouts to reduce compounding error in model-generated rollouts.
In the policy optimization part, we use decentralizd PPO \cite{schulman2017proximal} with a localized extended value function. Theoretically, we prove that the policy gradient computed 
from extended value function is a close approximation to the true gradient.
Empirically, we evaluate our method on adaptive traffic signal control (ATSC) and  connected autonomous vehicle (CAVs) control tasks \cite{chu2020multiagent, vinitsky2018benchmarks}, which are extensively studied intelligent transportation systems.

In summary, our contributions are three-fold. 
Firstly, we propose an algorithmic framework, $\ourmethod$,  for decentralized model-based reinforcement learning for networked systems.
Secondly, we integrate branched rollout to reduce compounding error in model-based rollouts and extended value function to reduce the complexity of computing policy gradient. We theorize that the resulting policy gradient is a close approximation to true policy gradient.
 Lastly, extensive results on intelligent transportation tasks demonstrate the superiority of $\ourmethod$ in terms of sample efficiency and performance.


\section{Related Work}
Reinforcement learning has achieved remarkable success in many decision making tasks \cite{mnih2015human,vinyals2019alphastar,silver2018general,han2019grid}. Due to low data efficiency, model-based methods are widely studied as a promising approach for improving sample efficiency \cite{reinforcementlearning,deisenroth2011pilco,luo2019algorithmic,janner2019trust}.

Due to the growing need of multi-agent decision making, such as traffic light control, wireless communications, and multi-player video games, many efforts have been poured in designing MARL algorithms. One line of work focuses on 
 centralized training decentralized execution (CTDE) framework, including policy gradients methods COMA \cite{foerster2018COMA}, MADDPG \cite{lowe2017MADDPG} and LIIR \cite{du2019learning}, and value factorization methods VDN \cite{sunehag2018value}, QMIX \cite{rashid2018qmix},  QTRAN \cite{son2019qtran}, etc. 
In large scale multi-agent systems, however, centralized training might not scale \cite{han2019grid}, and fully decentralized algorithms  are favoured. \cite{zhang2018fully} proposed an algorithm of NSC that can be proven to converge under linear approximation. \cite{qu2020scalable} proposed truncated policy gradient, to optimize local policies with limited communication. Baking in the idea of truncated $Q$-learning in \cite{qu2020scalable}, we generalize their algorithm to deep RL, rather than tabular RL. Factoring environmental transition into marginal transitions can be seen as factored MDP. \cite{guestrin2001multiagent} used Dynamic Bayesian Network to predict system transition. \cite{simao2019safe} proposed a tabular RL algorithm to ensure policy improvement at each step. However, our algorithm is a deep RL algorithm, enabling better performance in general tasks. {\cite{ruan2022GCSgraph, du2021flowcomm,foerster2016dial, zhang2013coordinating, NIPS2016_55b1927f} communicate with other agents by learning some hidden information. In comparison, our algorithm only needs to obtain the state of the neighbors.}

Early attempts on model-based MARL learning are restricted to special settings. 
For example, \cite{BRAFMAN200031} solved single-controller-stochastic games, which is a certain type of two-player zero-sum game;  \cite{zhang2020model} proved that model-based method can be nearly optimally sample efficient in two-player zero-sum Markov games; \cite{DBLP:conf/ijcai/0001WSZ21} constructs dynamics and opponents model in a decentralized manner, but it assumes full observability of states and the scalability is not sufficiently verified.
 \cite{bargiacchi2020model} extended model-based prioritized sweeping into a MARL scenario, but only restricted to tabular reinforcement algorithm, thus unable to deal with more general Markov decision making tasks. 

In contrast to existing works, this work tackles the more general multi-agent MDP under partial observability, and proposes the first fully decentralized model-based reinforcement learning algorithm.




\section{Problem Setup}

In this section, we introduce multi-agent networked MDP and model-based networked system control.
\subsection{Networked MDP}
We consider environments with a graph structure. Specifically, $n$ agents coexist in an underlying undirected and stationary graph $\mathcal{G}=(\mathcal{V},\mathcal{E})$. Agents are represented as a node in the graph, therefore $\mathcal{V}=\{1,...,n\}$ is the set of agents.
$\mathcal{E}\subset\mathcal{V}\times\mathcal{V}$ comprises the edges that represent the connectivity of agents. Agents are able to communicate along the edges with their neighbors. Let $N_i$ denote the neighbor of the agent $i$ including $i$ itself.
Let $N^\kappa_i$ denote the $\kappa$-hop neighborhood of $i$, i.e. the nodes whose graph distance to $i$ is less than or equal to $\kappa$. For the simplicity of notation, we also define $N_{-i}^\kappa=\mathcal{V}\setminus N_i^\kappa$. 

The corresponding networked MDP is defined as $(\mathcal{G}, \{\mathcal{S}_i, \mathcal{A}_i\}_{i\in\mathcal{V}},p,r)$. 
Each agent $i$ has its local state $s_i\in\mathcal{S}_i$, and performs action $a_i\in\mathcal{A}_i$. The global state is the concatenation of all local states: $s=(s_1,...,s_n)\in\mathcal{S}:=\mathcal{S}_1\times ...\times\mathcal{S}_n$. Similarly, the global action is  $a=(a_1,...,a_n)\in\mathcal{A}:=\mathcal{A}_1\times...\times\mathcal{A}_n$.
For the simplicity of notation, we define $s_{N_i}$ to be the  states of $i$'s neighbors.
The transition function is defined as: $p(s'|s,a): \mathcal{S}\times\mathcal{A}\rightarrow \mathcal{S}$.
Each agent possess a localized policy $\pi_i^{\theta_i}(a_i|s_{{N_i}})$ that is parameterized by $\theta_i\in\Theta_i$, meaning the local policy is dependent only on states of its neighbors and itself. We use $\theta=(\theta_1,...,\theta_n)$ to denote the tuple of localized policy parameters, and $\pi^\theta(a|s)=\prod_{i=1}^n\pi_i^{\theta_i}(a_i|s_{{N_i}})$ denote the joint policy.
{Each agent has a reward functions that depends on local state and action: $r_i(s_i, a_i)$,} and the global reward function is defined to be the average reward $r(s,a)=\frac{1}{n}\sum_{i=1}^nr_i(s_i,a_i)$. 
The goal of reinforcement learning is to find a policy $\pi^{\theta}$ that maximizes the discounted reward, \begin{equation}
\pi^{\theta^\ast} 
=\arg\max_{\pi^\theta}\mathbb{E}_{\pi^{\theta}}\Big[\sum_{t=0}^\infty\gamma^t r(s_t,a_t)\Big],
\end{equation}
where $\gamma\in(0,1)$ is the temporal discount factor. 
The value function is defined as 
\begin{align}
    V(s)=\mathbb{E}_{\pi^\theta}\Big[\sum_{t=0}^{\infty}\gamma^t r(s_t,a_t)|s^0=s\Big]=\frac{1}{n}\sum_{i=1}^nV_i(s). 
\end{align}
In the last step, we have defined $V_i(s)$, which is the value function for individual reward $r_i$.


\subsection{Model-based RL}

Let $V^{\pi,  \widehat{p}}$ be the value function of the policy on the estimated model $\widehat{p}$. 
Towards optimizing $V^{\pi,p}(s)$, a common solution in single agent RL is to build a lower bound as follows and maximize it iteratively \cite{luo2019algorithmic}:
\begin{equation}\label{eq:lowerbound}
V^{\pi, p}(s) \geq V^{\pi, \widehat{p}}(s)  - D( \widehat{p}_i, \pi_i),
\end{equation}
where $D( \widehat{p}, \pi) \in \mathbb{R}$ bounds the discrepancy between $V_i^{\pi, p}$ and $V_i^{\pi, \widehat{p}}$ and can be defined as $D( \widehat{p}_i, \pi_i) = \alpha \cdot \mathbb{E}[\|  \widehat{s}_{i,t+1}- s_{i,t+1}\|]$,
where $\alpha$ is a hyperparameter.

\begin{figure*}[th!]
	\centering
	 	\includegraphics[width=0.9\linewidth]{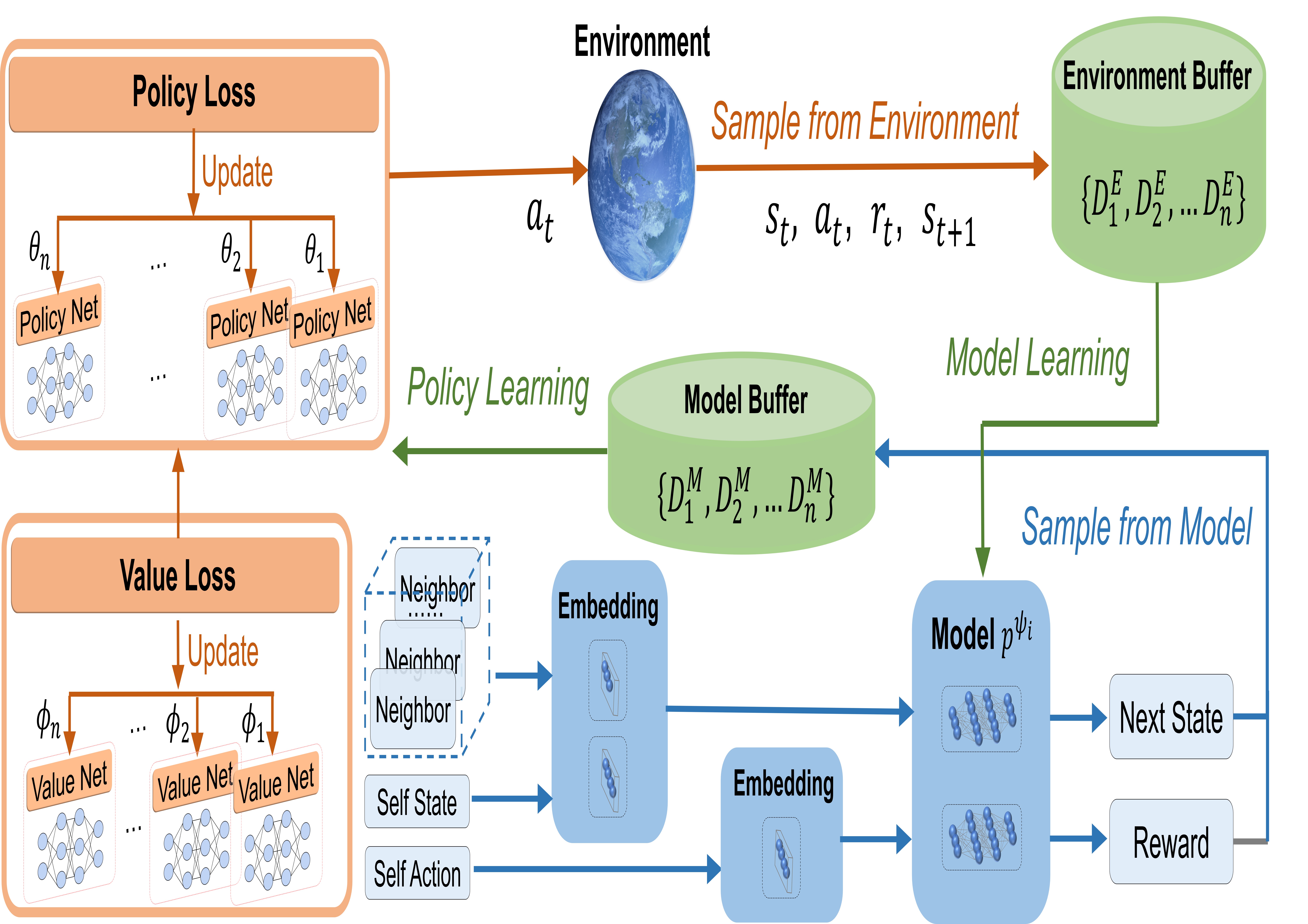}
    \caption{ Architecture of $\ourmethod$. Experience data of state transition and rewards from interaction between the environment and the agent are used to train the environment model. Then $\ourmethod$ can use the large amount of data generated by the interaction between the model and the agent to improve the sampling efficiency in the training process. 
    }
\label{fig:arc-dmpo}
\vspace{-0.2 in}
\end{figure*}

\begin{algorithm}[t!]
    \caption{$\ourmethod$ framework}
    \label{alg:DMPO}
    \KwIn{rollout length $T$}
    \begin{algorithmic}[1]
    \STATE {Initialize the model $p^{\psi_i}$, actor $\pi^{\theta_i}$ and critic $V^{\phi_i}$.}
    \STATE {Initialize replay buffers $\mathcal{D}^{E}_i$ and $\mathcal{D}^{M}_i$.}
    \FOR{$N$ epochs}
        \STATE Take action in environment according to $\pi^{\theta_i}, i\in \mathcal{V}$; add to $D^E_i, i\in\mathcal{V}$
            \STATE{Train $p^{\psi_i}$ on $D^{E}_i$ by maximizing likelihood, $i\in\mathcal{V}$.}
        \STATE{$D^{model}_i = \emptyset$.} 
        \FOR{$B$ steps}  
                \FOR{$M$ rollouts}
                    \STATE Sample $s_i^t$ from $D^E_i$, $i\in\mathcal{V}$ \tcp{branching}
                    \STATE Generate $T$-step rollout initing from $\{s^t_i\}_{i\in\mathcal{V}}$ by policy $\pi^{\theta_i}$ and model ${p}^{\psi_i}$, $i\in\mathcal{V}$
                    \STATE Append trajectories to $D^M$
                \ENDFOR  
            \FOR{$G$ gradient steps}
                \STATE
                    Update policies $\theta_i$ and critics $\phi_i$ on model data sampled from $D^M$
            \ENDFOR
        \ENDFOR
    \ENDFOR
    \end{algorithmic}
\end{algorithm}

\section{Decentralized Model-based Policy Optimization} \label{sec:alg}


In this section, we formally present $\ourmethod$, which is a decentralized model-based reinforcement learning algorithm. Three key components are localized models, decentralized policies and extended value functions. 

\subsection{Modeling Networked System}
Networked system may have some extent of locality, meaning in some cases, local states and actions do not affect the states of distant agents. In such systems, environmental transitions can be factorized, and agents are able to maintain local models to predict future local states. The factorization is given below.
\begin{align}\label{eq:factorization-nsc}
 {p}(s,a)=\prod_{i=1}^n {p}_i(s_i'|s_{{N_i^{\kappa}}},a_i).
\end{align}
While ${N_i}$ is often not known, we employ the $\kappa$-neighbor $N_i^{\kappa}$ to approximate the true dynamic by $ \widehat{p}(s,a)= \prod_{i=1}^n  \widehat{p}_i(s_i'|s_{{N_i^{\kappa}}},a_i)$, where $ \widehat{p}_i$ is usually parameterized by unknown variables that we denote as $\psi_i, i\in \mathcal{V}$.
Larger $\kappa$ leads to better approximation of model $p$, but also more computation overhead.
In the following presentation, we would use $ \widehat{p}_i$ and  ${p}^{\psi_i}$ interchangeably.

For each agent $i$, we solve the following problem:
\begin{equation}\label{eq:opt-general}
\begin{aligned}
\pi^{k+1}_i, p^{k+1}_i = \underset{\pi_i, p_i}{\operatorname{argmax}} \quad V^{\pi, p}-D( \widehat{p}_i, \pi_i). 
\end{aligned}
\end{equation}
Let $ \widehat{s}_{i,t+1} = {p}^{\psi_i}({s}_{N_i,t},a_i)$. We want to minimize$\| \widehat{s}_{i,t+1} - s_{i,t+1}\|$.
With trajectories sampled from the true environment by $\pi_i$, each agent locally updates its model $ \widehat{p}_i$. 
With a localized model $ \widehat{p}_i$, agent $i$ learns to update $\pi_i$ to maximize the reward. 

For the training of policies and models, we maintain two data buffers, $\mathcal{D}^{E}$ for trajectories generated by true environment $p$ and $\mathcal{D}^{M}$ for data generated by the learned model. \textcolor{black}{The trajectories here consist of $s$, $a$, $s'$, $r$, $d$, where $d$ is binary, indicating whether the task is completed or reaches maximum episode length.}
The architecture of our framework is presented in Figure \ref{fig:arc-dmpo}.
Below we present the details of model and policy updates.

\subsection{Update policies}\label{subsec:learning}

To optimize the policies, we need to adopt an algorithm that can exploit network structure. Whilst remaining decentralized. Independent RL algorithms that observe only local states are fully decentralized, but they often fail to learn an optimal policy. Centralized algorithms that utilize centralized critics often achieve better performance than decentralized algorithms, but they might not scale to large environments where communication costs are expensive. 

For each agent $i$, denote parameterized policy $\pi^{\theta_i}$ and critic $V^{\phi_i}$ for fitting optimal policy $\pi_i^*$ and critic $V_i(s)$.
Let $\{s_{i,\tau}, a_{i,\tau}, r_{i,\tau}\}_{i \in \mathcal{V}, \tau\in \mathcal{B}}$ a minibatch sample from $D^M$ under policies $\pi^{\theta_i}, i\in \mathcal{V}$.
Define the advantage function
$\hat{A}^{(t)}=r^{(t)}+\gamma V(s^{(t+1)})-V(s^{(t)})$.
As it is not easy to obtain the true value $V_i(s)$, we adopt an \textit{extended value function}, which is defined as 
\begin{align}
    V_i(s_{N_i^\kappa})=\mathbb{E}_{s_{N_{-i}^\kappa}}\bigg[\sum_{t=0}^\infty r_i^t|s^0_{N_i^\kappa}=s_{N_i^\kappa}\bigg],i\in\mathcal{V}.
\end{align} 
Note that $V_i(s_{N_i^\kappa})$ is a good approximation of $V_i(s)$, with the discrepancy decreasing exponentially with $\kappa$. We defer the discussion to Section \ref{sec:extendedvalue}.

To generate the objective for extended value function, or return $R_i$, we use reward-to-go technique. However, because model rollout is short, standard reward-to-go returns would get a biased estimation of $V_i(s)$, a.k.a. compounding error of model generation. To resolve this issue, we add the value estimation of the last state to the return. 
The target of $V_i(s^t_{N_i^\kappa})$ is
\begin{equation} \label{eq:returns}
R_i^t = \sum_{l=0}^{T-t-1}\gamma^l r_i^{t+l} + V^{\phi_i}(s_{N_i^\kappa}^{T}).
\end{equation}
The loss of value function is defined as 
\begin{align}
 \mathcal{L}(\psi_i)=\frac{1}{|\mathcal{B}|}\sum_{\tau\in \mathcal{B}}\big(V^{\phi_i}(s_{N_i^\kappa}^{\tau})-R_i^{\tau}\big)^2.
\end{align}
Empirically, we make use of communication within $\kappa$-hop neighbors and  generate an estimation of global value function as
\begin{equation} \label{eq:value}
    \tilde{V}_i (s^t_{N_i^\kappa})=\frac{1}{n}\sum_{j\in N_i^\kappa}V_j(s^t_{N_j^{\kappa}})
\end{equation}
The advantage is thus defined as $\hat{A}_t=r_i^{t}+\gamma \tilde{V}_i (s^{t+1}_{N_i^\kappa})- \tilde{V}_i (s^t_{N_i^\kappa}))$. 
The loss function of a $\ourmethod$ agent is defined as
\begin{equation} \label{eq:policy_loss}\small
    \mathcal{L}(\theta_i) = \frac{1}{|\mathcal{B}|}\sum_{\tau\in \mathcal{B}}\big(-\log\pi^{\theta_i}(a_{i,\tau}|s_{N_i,\tau})\hat{A}_{i,\tau} + \beta H(\pi^{\theta_i})\big).
\end{equation}
{We replace $\log\pi^{\theta}$ with $\frac{\pi^{\theta}}{\pi^{\theta_{\text{old}}}}$ and adopt a PPO agent \cite{schulman2017proximal} to constrain the policy shift in implementations.}

In general, larger $\kappa$ leads to better results on  model approximation and policy learning, but also more communication costs. In algorithmic solutions, computation cost cannot be ignored as larger $\kappa$ leads to more complex models or policies architectures, posing difficulties on training. We discuss more about  this in experiments. 

\subsection{Update Model}\label{subsec:model}

To perform decentralized model-based learning, we let each agent maintain a localized model. The localized model can observe the state of $\kappa$-hop neighbor and the action of itself, and the goal of a localized model is to predict the information of the next timestep, including state, reward. This process is denoted by ${p}^{\psi_i}(s_i', r_i'|s_{{N_i^{\kappa}}},a_i)$.
In practice, the data are all stored locally by each agent. We minimize the following objective to update models.
\begin{align}
    \mathcal{L}(\psi_i) = \frac{1}{|\mathcal{B}|}\sum_{\tau\in \mathcal{B}} \| \widehat{s}_{i,\tau+1} - s_{i,\tau+1}\|^2.
\end{align}

Scaling model-based methods into real tasks can result in decreased performance, even if the model is relatively accurate. One main reason is the compound modeling error when long model rollouts are used, and model error compounds along the rollout trajectory, making the trajectory ultimately inaccurate. To reduce the negative effect of model error, we adopt a branched rollout scheme proposed in \cite{janner2019trust,sutton1990integrated}. In branched rollout, model rollout starts not from an initial state, but from a state that is randomly selected from the most recent environmental trajectory $\tau$. Additionally, the model rollout length is fixed to be $T$, as indicated in line 10 in Algorithm \ref{alg:DMPO}. Line 4-5 of Algorithm \ref{alg:DMPO} describe the model training steps.



\section{Theoretical Analysis}\label{sec:extendedvalue}
In this section, we discuss that the extended value function $V_i(s_{N_i^\kappa})$ is a good approximation of the real value function. 
We formally state the result in Theorem \ref{thm:V} and defer the proof to Appendix.
\begin{mythm}\label{thm:V}
Define $V_i(s_{N_i^\kappa})=\mathbb{E}_{s_{N_{-i}^\kappa}}[\sum_{t=0}^\infty r_i^t(s_t,a_t)|s^0_{N_i^\kappa}=s_{N_i^\kappa}]$, and $V_i(s)=\mathbb{E}[\sum_{t=0}^\infty r_i^t|s^0=s]$, then
\begin{equation}\label{thm1:eqV}
    |V_i(s)-V_i(s_{N_i^\kappa})|\leq \frac{r_{\max}}{1-\gamma}\gamma^\kappa.
\end{equation}
\end{mythm}
\begin{myremark}
Recall that $V(s) =\frac{1}{n}\sum_{i=1}^nV_i(s)$. From Eq. \eqref{thm1:eqV}, it is easy to obtain the following result,
\begin{equation}
    |V(s)-\frac{1}{n}\sum_{i=1}^n V_i(s_{N_i^\kappa})| \leq \frac{r_{\max}}{1-\gamma}\gamma^\kappa,
\end{equation}
{which indicates that the global value function $V(s)$ can be approximated by the average of localized value functions.}
\end{myremark}

In policy optimization, value functions are used for calculating advantages $\hat{A}^{(t)}$, and we have shown that $V(s)$ can be estimated with the average of 
localized
value functions $\frac{1}{n}\sum_{i=1}^n V_i(s_{N_i^\kappa})$. In practice, an agent might not get the value function of distant agents and can only access the value function of its $\kappa$-hop neighbors. However, we can prove that $\tilde{V}_i=\frac{1}{n}\sum_{j\in N_i^\kappa}V_j(s_{N_j^\kappa})$ is already very accurate for calculating the policy gradient for agent $i$.
Theorem \ref{thm:pg} formally states this result and the proof is deferred to Appendix.
\begin{mythm}\label{thm:pg}
Let $\hat{A}_t=r^{(t)}+\gamma V(s^{(t+1)})-V(s^{(t)})$ be the TD residual, and
$g_i=\mathbb{E}[\hat{A}\nabla_{\theta_i}\log\pi_i(a|s)]$ be the policy gradient. If $\tilde{A}_t$ and $\tilde{g}_i$ are the TD residual and policy gradient when value function $V(s)$ is replaced by $\tilde{V}_i(s)=\frac{1}{n}\sum_{j\in N_i^\kappa}V_j(s_{N_i^\kappa})$, we have:
\begin{equation}\label{thm2:graddiff}
    |g_i-\tilde{g}_i| \leq \frac{\gamma^{\kappa-1}}{1-\gamma}[1-(1-\gamma^2)\frac{N_i^\kappa}{n}]r_{\max}g_{\max},
\end{equation}
where $r_{\max}$ and $g_{\max}$ denote the upper bound of the absolute value of reward and gradient, respectively.
\end{mythm}
\begin{myremark}
Theorem \ref{thm:pg} justifies that the policy gradients computed based on the sum of the neighboring extended value functions is a close approximation of true policy gradients. 
The power of this theorem is that the extended value function $V_i(s_{N_i}^{\kappa})$ requires only the neighboring information, thus easier to approximate and scalable. Despite the reduction in computation, the difference between the approximated and true gradient in Eq. \eqref{thm2:graddiff} is small.

\end{myremark}

\begin{figure}[h!]
	\centering

    \subfigure[Ring]{
    		\includegraphics[width=0.32 \columnwidth]{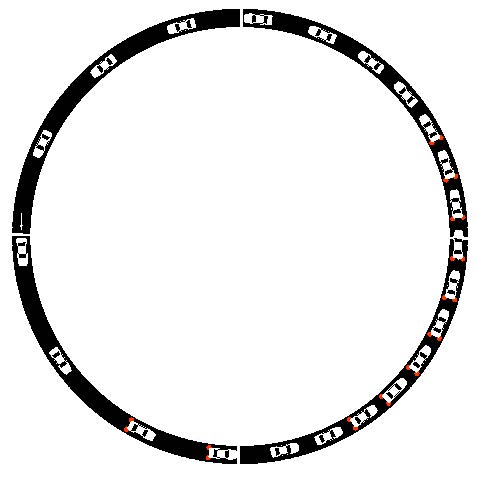}
    		}
    \subfigure[Figure Eight]{
    		\includegraphics[width=0.32 \columnwidth]{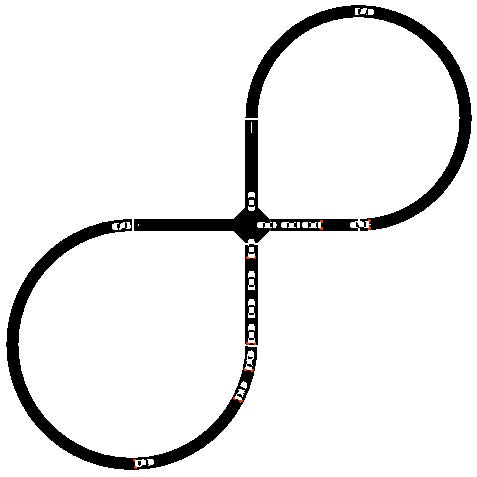}
    		}
	\subfigure[CACC]{
    		\includegraphics[width=0.35 \columnwidth]{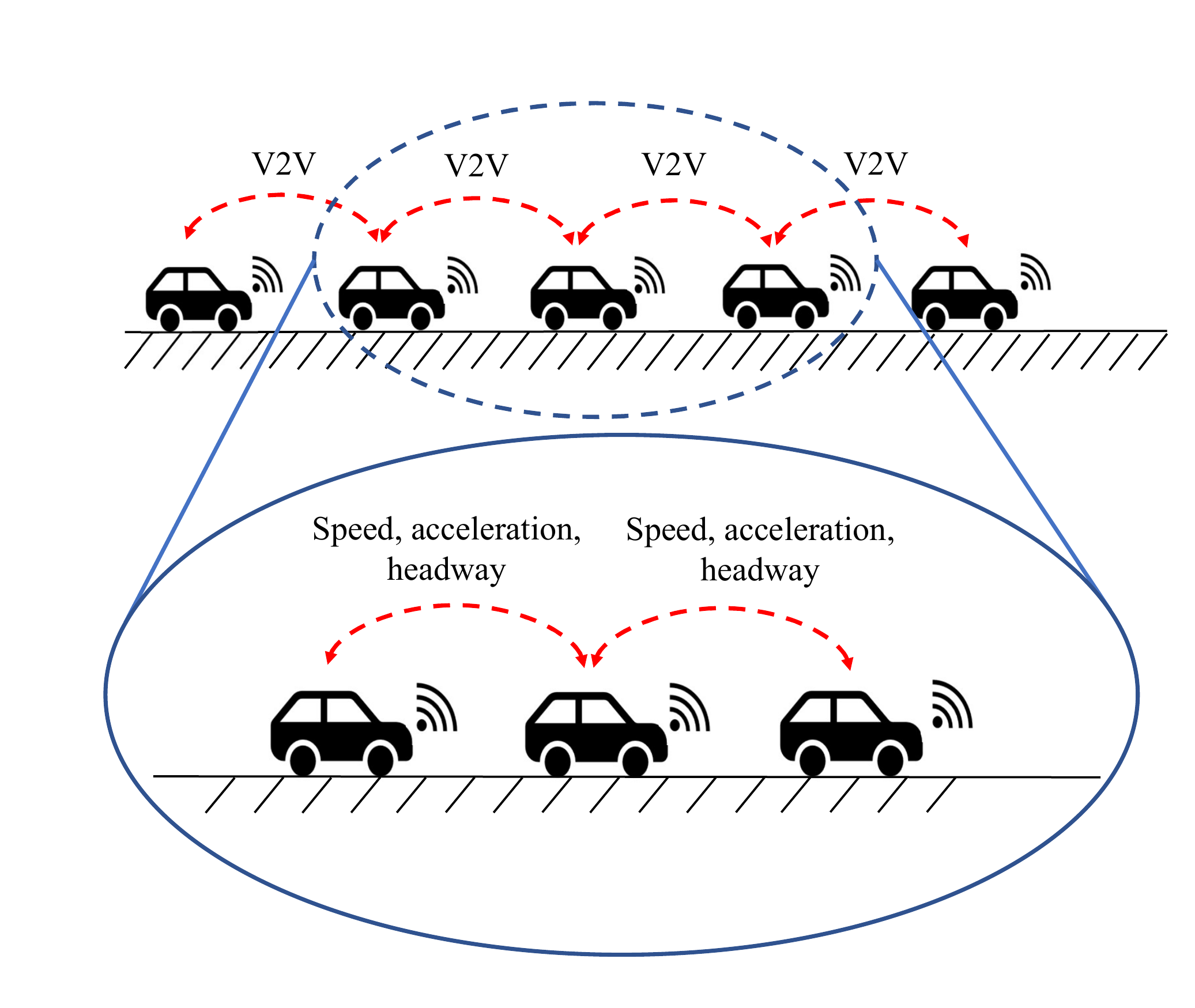}
    		}
    \subfigure[ATSC Grid]{
    		\includegraphics[width=0.32 \columnwidth]{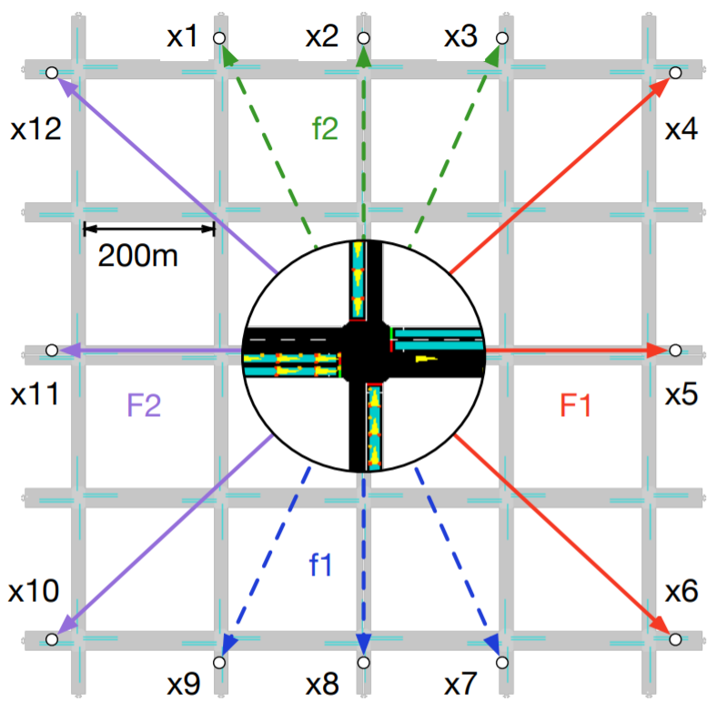}
    		}
    \caption{Visualization of  CACC, Flow and ATSC environments. (a) Vehicles travel in a ring to reduce stop-and-go waves. (b) Vehicles travel in a figure eight shaped road section to learn the behavior at an intersection. (c) A line of vehicles that need to keep a stable velocity and desired headway. (d) Synthetic traffic grid that need to learn how to minimize traffic congestion. 
    } 
    \label{fig:task-visulization}
\end{figure}

\begin{table}[t!]
    \centering   
    \caption{Hyperparameters for DMPO.}
    \begin{tabular}{c|c|c|c|c|c}
        \hline
         & Catch. & Slow. & Fig.Eight & RingAtt. & ATSC\\
        \hline
        lr of $V_i$ & 3e-4 & 3e-4 & 5e-5 & 5e-4 & 5e-4 \\
        \hline
        lr of $\pi$ & 3e-4 & 3e-4 & 5e-5 & 5e-4 & 5e-4 \\
        \hline
        lr of $p_i$ & 3e-4 & 3e-4 & 5e-4 & 5e-4 & 2e-4 \\
       
        \hline
        $\pi_i$ Net. & [64,64] & [64,64] & [64,64] & [64,64] & [128,128] \\
        \hline
        $V_i$ Net. & [64,64] & [64,64] & [64,64] & [64,64] & [128,128] \\
        \hline
        $p_i$ Net. & [16,16] & [16,16] & [16,16] & [16,16] & [64,64] \\
        \hline
        $\kappa$ & 2 & 2 & 3 & 3 & 1 \\
         \hline
        Rollout  & 25 & 25 & 25 & 25 & 25 \\
        \hline
    \end{tabular}
    \label{tab:param_dmpo}
\end{table}

\section{Experiments}


We evaluate $\ourmethod$ on existing multi-agent environments of intelligent transport systems, which are Connected Autonomous Vehicles including Flow \cite{vinitsky2018benchmarks} and  Cooperative Adaptive Cruise Control (CACC) \cite{chu2020multiagent},  and Adaptive Traffic Signal Control (ATSC) \cite{chu2020multiagent}. 
{The number of agents in these systems is 8, 14, 22 and 25 respectively, and the complexity of the networked systems gradually increases.} Figure \ref{fig:task-visulization} gives visualization for the environments used in the experiments. 


We use three-layered MLP layers for policy, critic and model prediction networks. For policy and critic, all hidden layers are set up to 64 hidden units for Flow and CACC, and 128 for ATSC. For model networks, all hidden layers are set to 16 hidden units for Flow and CACC, and 64 for ATSC.  $\kappa$ for policy and model are set up to 1 in all tasks. For critics, $\kappa$ is set to 2 for CACC, 3 for Flow and 1 for ATSC.
The key hyperparameters for $\ourmethod$ are summarized in Table \ref{tab:param_dmpo}.

\subsection{Baselines}
We evaluate the following algorithms in experiments.
\begin{itemize}
    \item CPPO \cite{schulman2017proximal,vinitsky2018benchmarks}: Centralized PPO learns a centralized critic $V_i(s)$. This baseline aims to analyze the performance when $\kappa$ is set to be arbitrarily huge, and is used in \cite{vinitsky2018benchmarks} as a benchmark algorithm for networked system control.

    \item DPPO \cite{schulman2017proximal}: Decentralized PPO learns an independent actor and critic for each agent. We implement it by using the neighbor's state for extended value estimation. 
    \item IC3Net \cite{singh2018learning}: A communication-based multi-agent RL algorithm. The agents maintain their local hidden states with a LSTM kernel, and actively determine the  communication target. Compared with DPPO, IC3Net employs communication, whereas DPPO agents only observe the states of their neighbors.

    \item DMPO (our method):  $\ourmethod$ is a decentralized and model-based algorithm based on DPPO. The extended value function is based on $\kappa$-hop neighbors.
    
\end{itemize}


\begin{figure}[t!]
	\centering
	\subfigure[Figure Eight]{
    		\includegraphics[width=0.44\columnwidth]{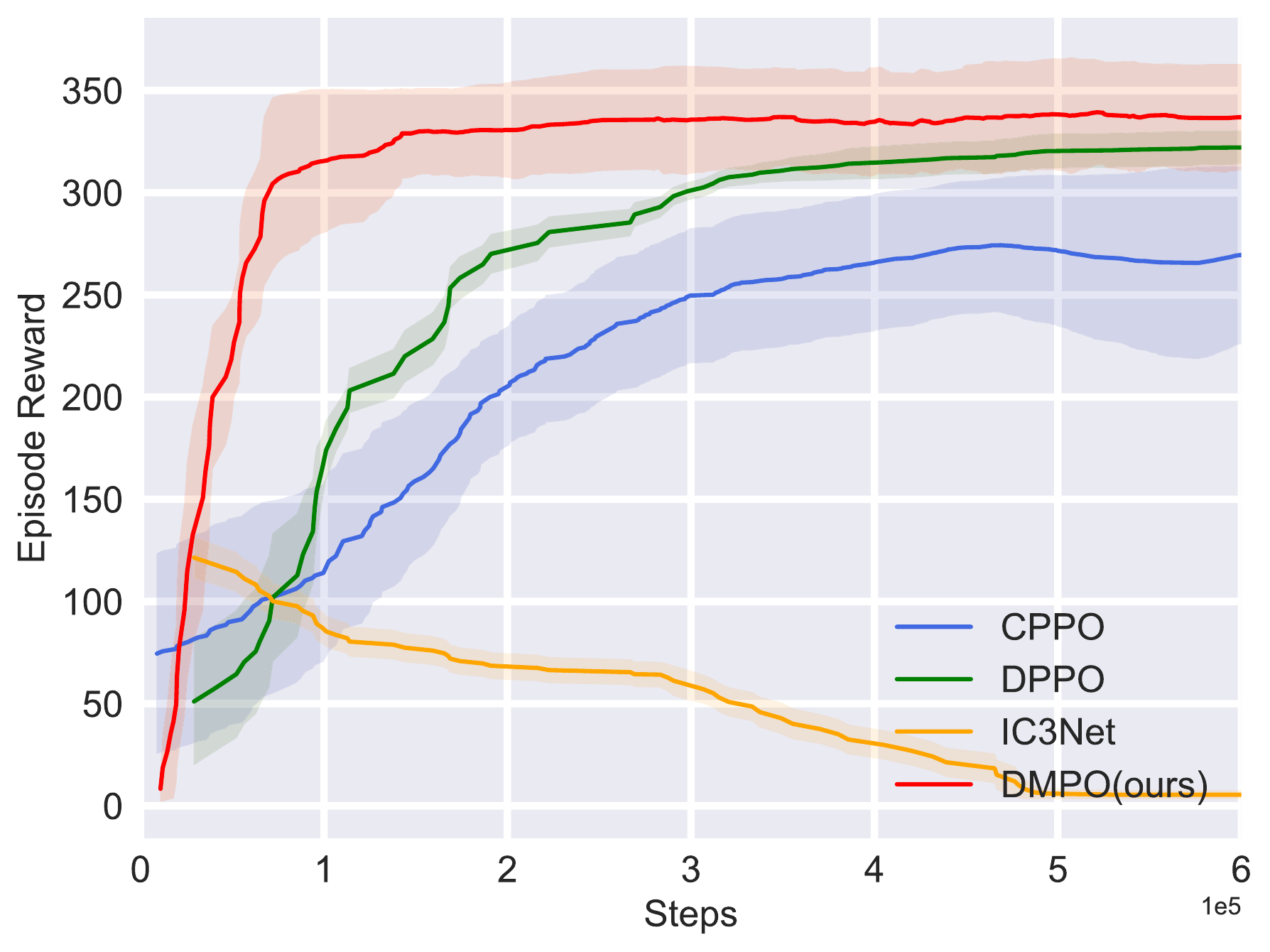}
    		\label{fig:eightres}
    		}
    \subfigure[Ring Attenuation]{
    		\includegraphics[width=0.46 \columnwidth]{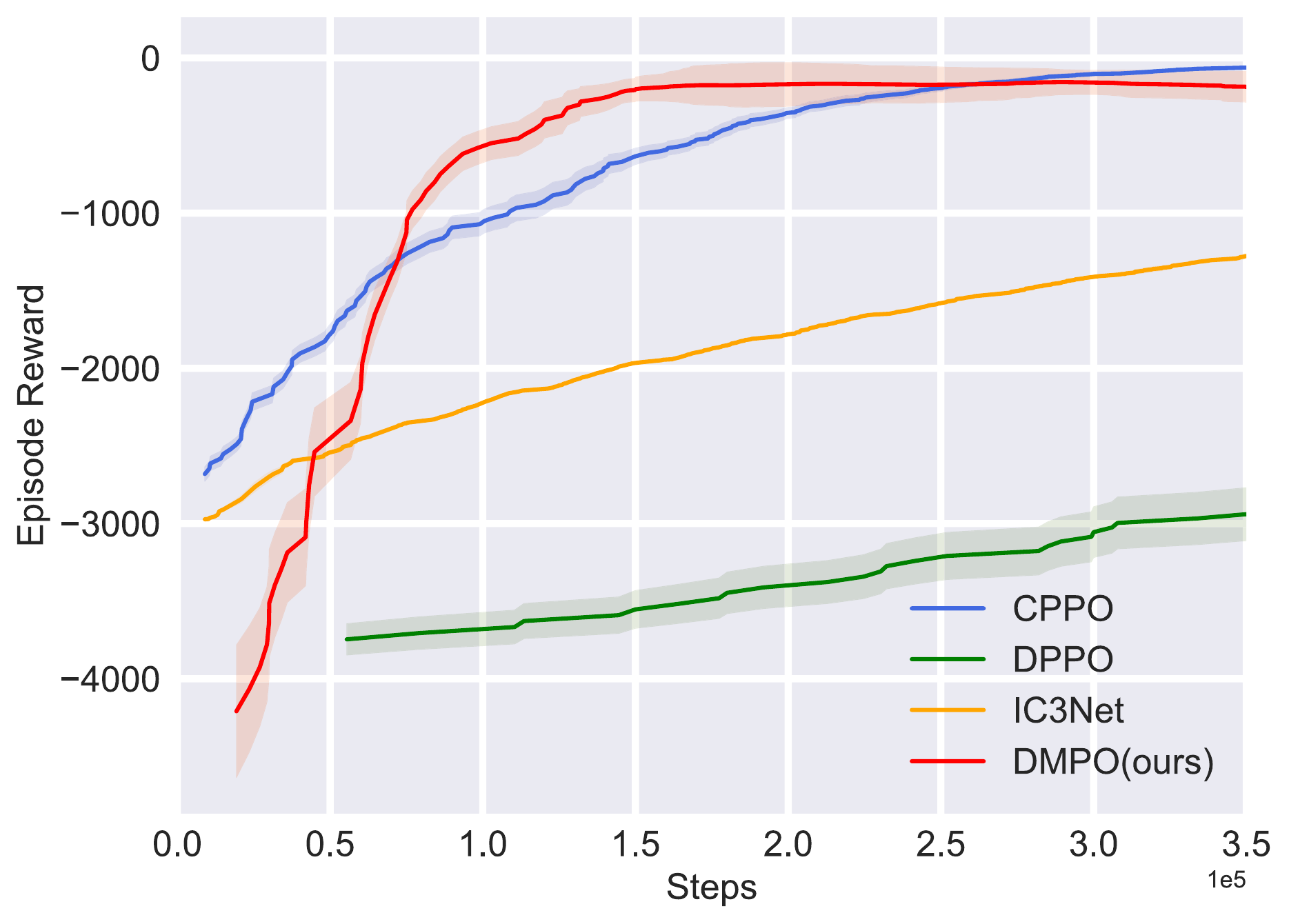}
    		\label{fig:ringres}
    		}
	\subfigure[CACC Catch-up]{
    		\includegraphics[width=0.46\columnwidth]{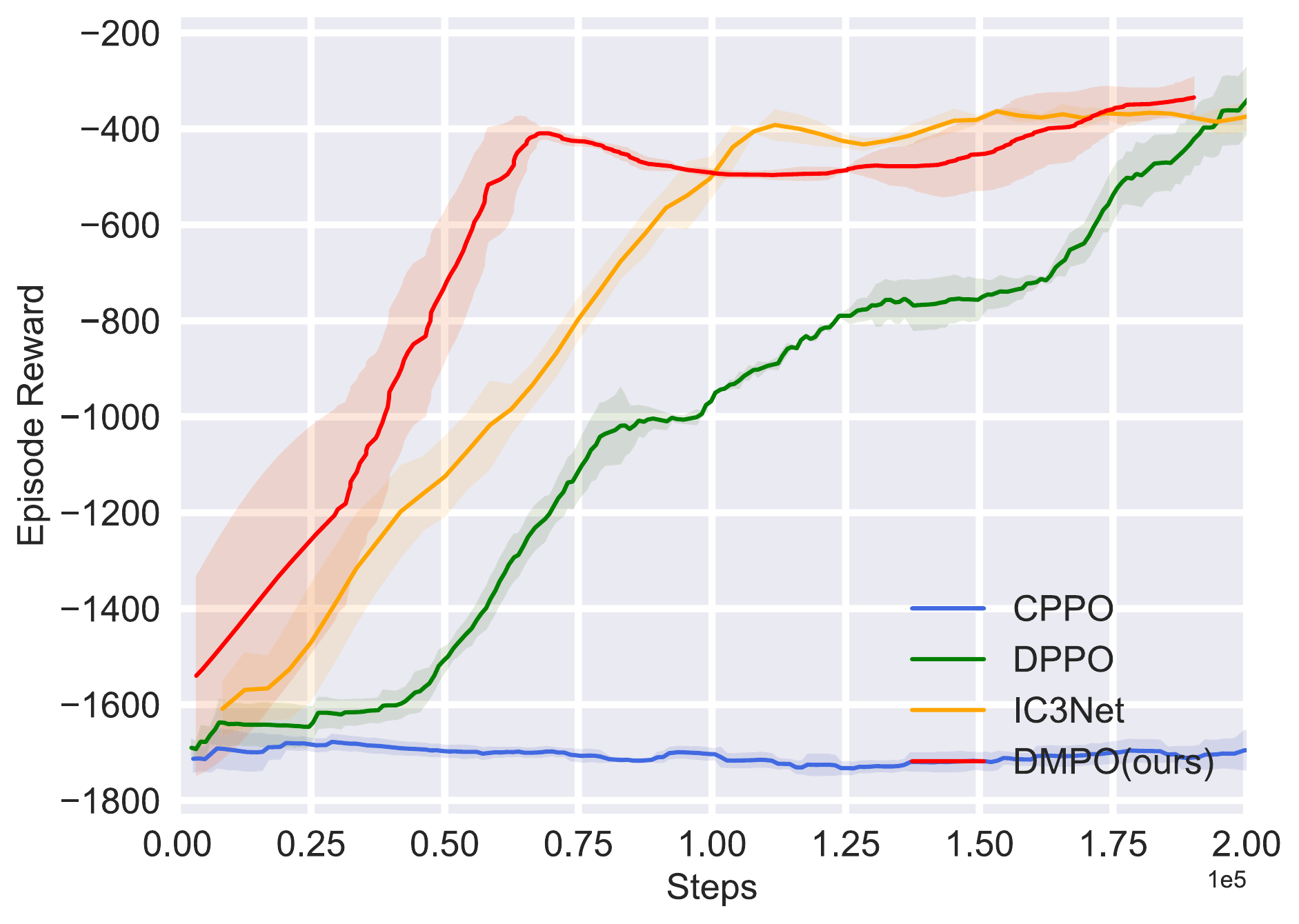}
    		\label{fig:catchupres}
    		}
    \subfigure[CACC Slow-down]{
    		\includegraphics[width=0.44 \columnwidth]{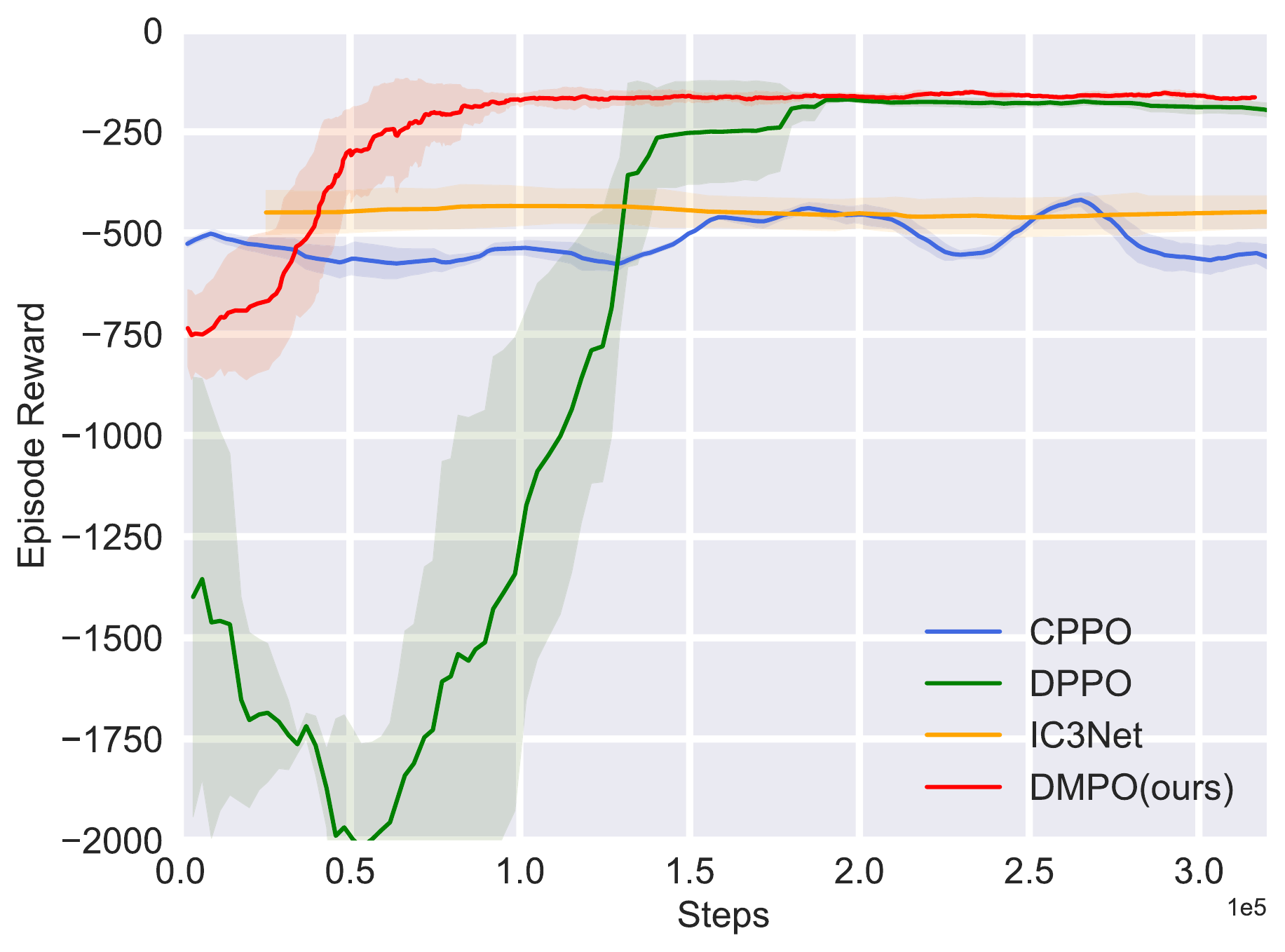}
    		\label{fig:slowdownres}
    		}
    		\label{fig:gridres}
    \caption{Training curves on multi-agent environments. Solid curves depict the mean of trails, and shaded region correspond to standard deviation.}
    \label{fig:CACC}
\end{figure}
\subsection{Connected Autonomous Vechicles}



{\textbf{Cooperative Adaptive Cruise Control }
CACC consists of two scenarios: Catch-up and Slow-down. The objective of CACC is to adaptively coordinate a platoon of 8 vehicles to minimize the car-following headway and speed perturbations based on real-time vehicle-to-vehicle communication.
{The state of each agent consists of headway $h$, velocity $v$, acceleration $a$ , and is shared to neighbors within two steps. The action of each agent is to choose appropriate hyper-parameters $\left(\alpha^{\circ}, \beta^{\circ}\right)$ for each OVM controller \cite{bando1995dynamical}}}

\textcolor{black}{\textbf{Flow environments} {This task consists of Figure Eight and Ring Attenuation. The objective is to achieve a target speed and avoid collision, which is similar to CACC. The road network has a shape of ring or figure ``eight".
The figure eight network, previously presented in \cite{wu2017emergent}, acts as a closed representation of an intersection. The state consists of velocity and position for the vehicle. The action is the acceleration of the vehicle.
In the perspective of a networked system, we assume that the vehicles are connected with the preceding and succeeding vehicle, thus resulting in a loop-structured graph.}}
\begin{figure}[t!]
	\centering
	\subfigure[Episode reward in Ring Att.  ]{
    		\includegraphics[width=0.46\columnwidth]{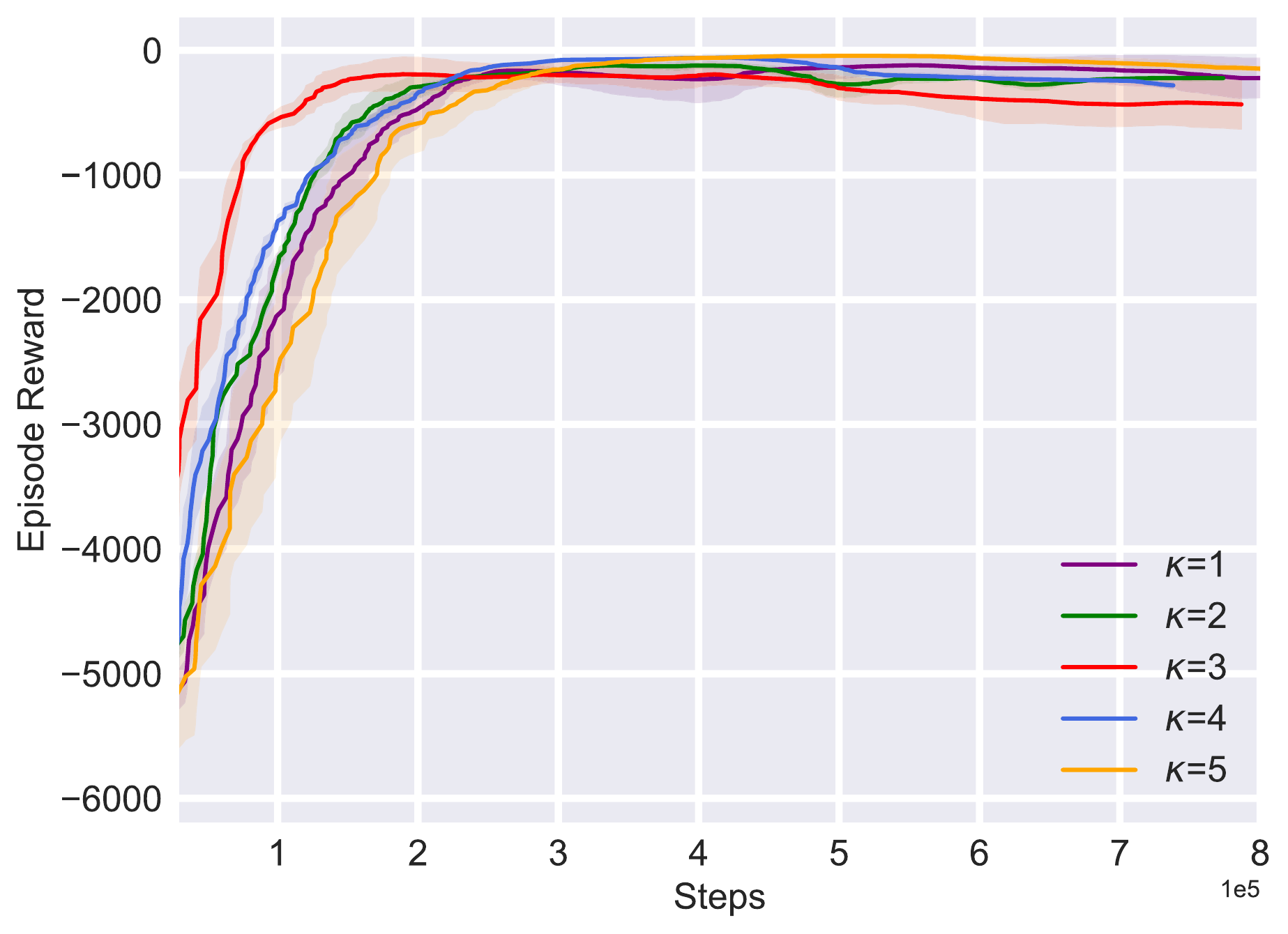}
    		}
	\subfigure[Model error in Ring Att. ]{
    		\includegraphics[width=0.44\columnwidth]{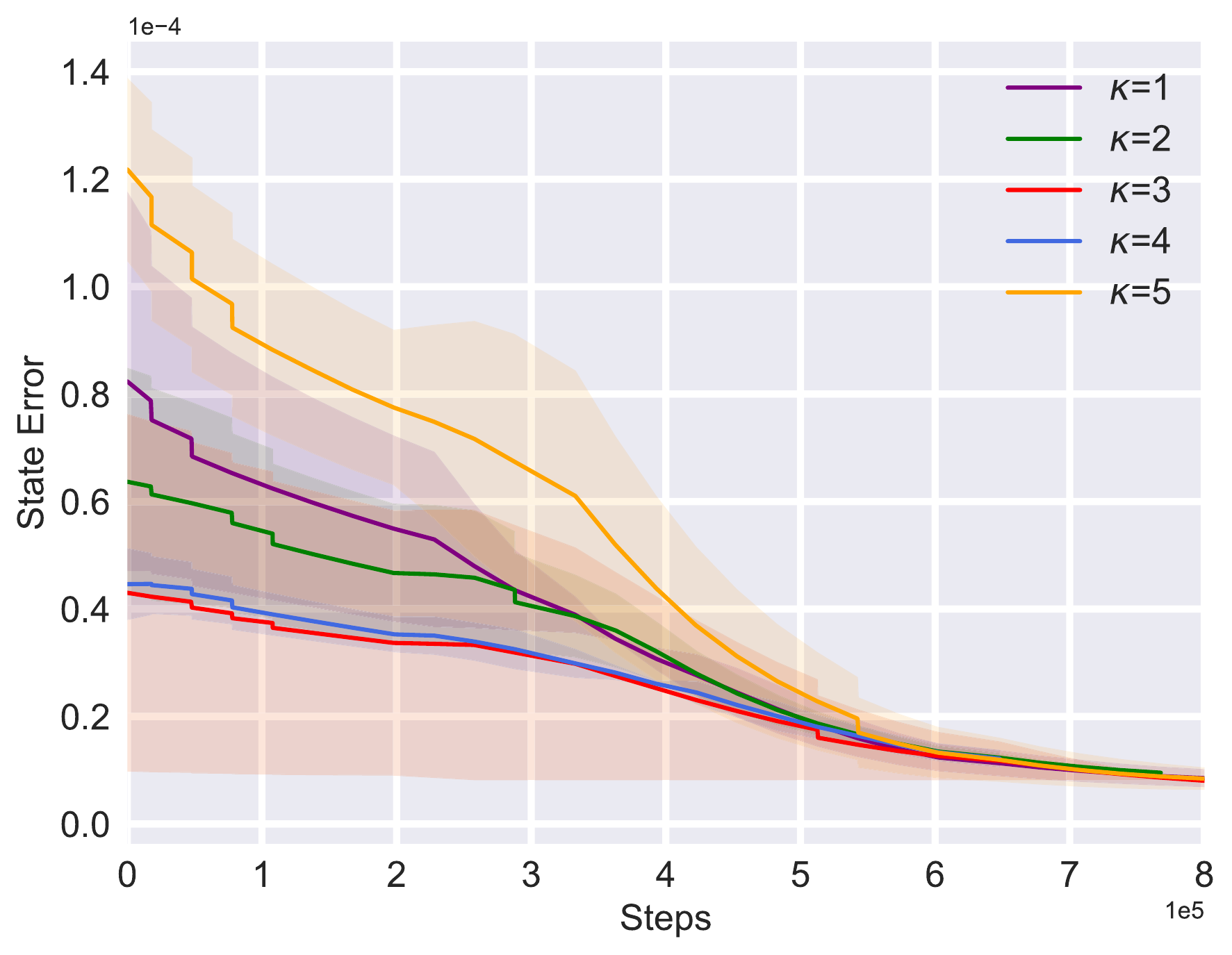}
    		}
    \subfigure[Episode reward in Catch-up]{
    		\includegraphics[width=0.46 \columnwidth]{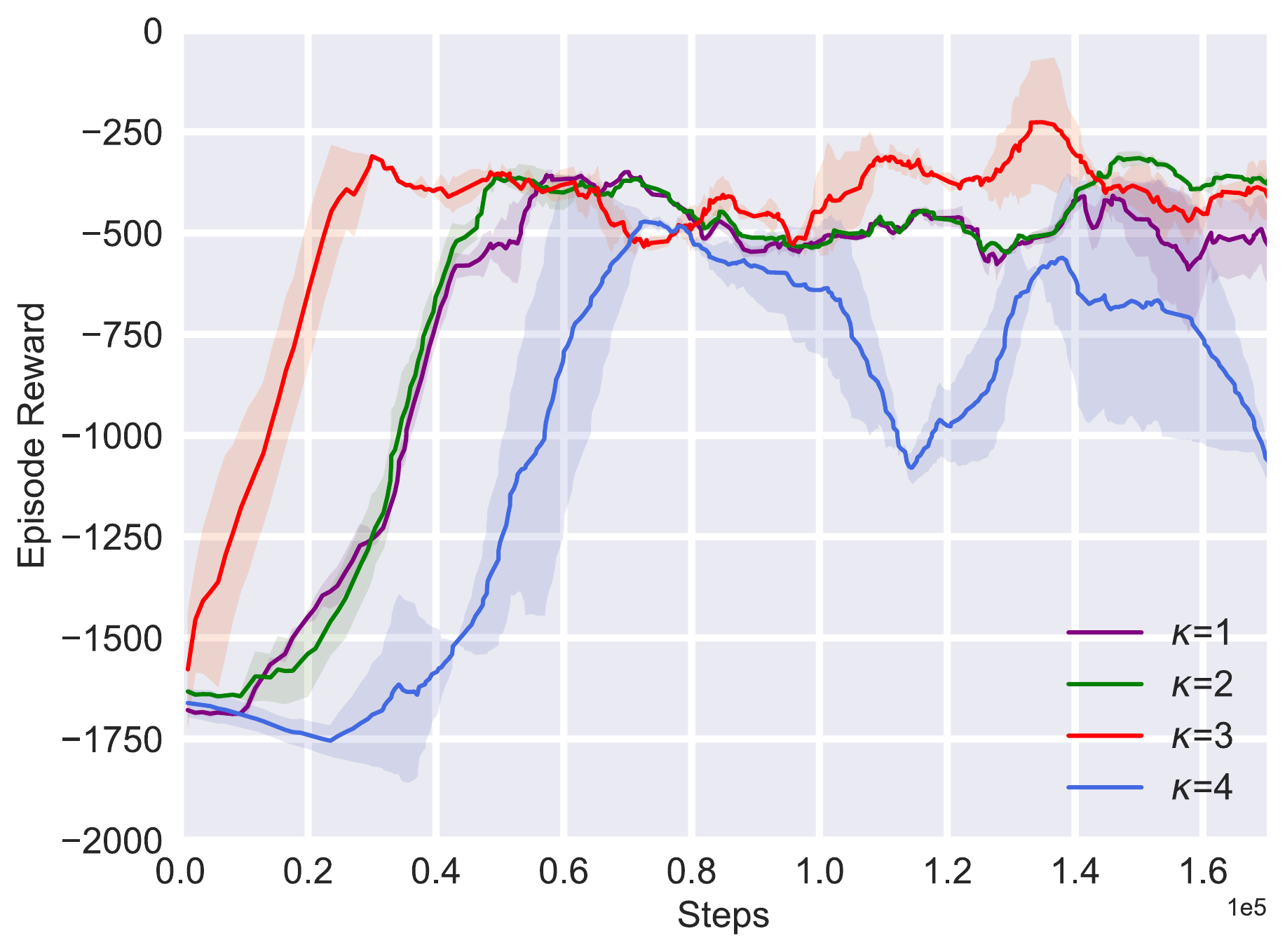}
    		}
    \subfigure[Model error in Catch-up ]{
    		\includegraphics[width=0.44 \columnwidth]{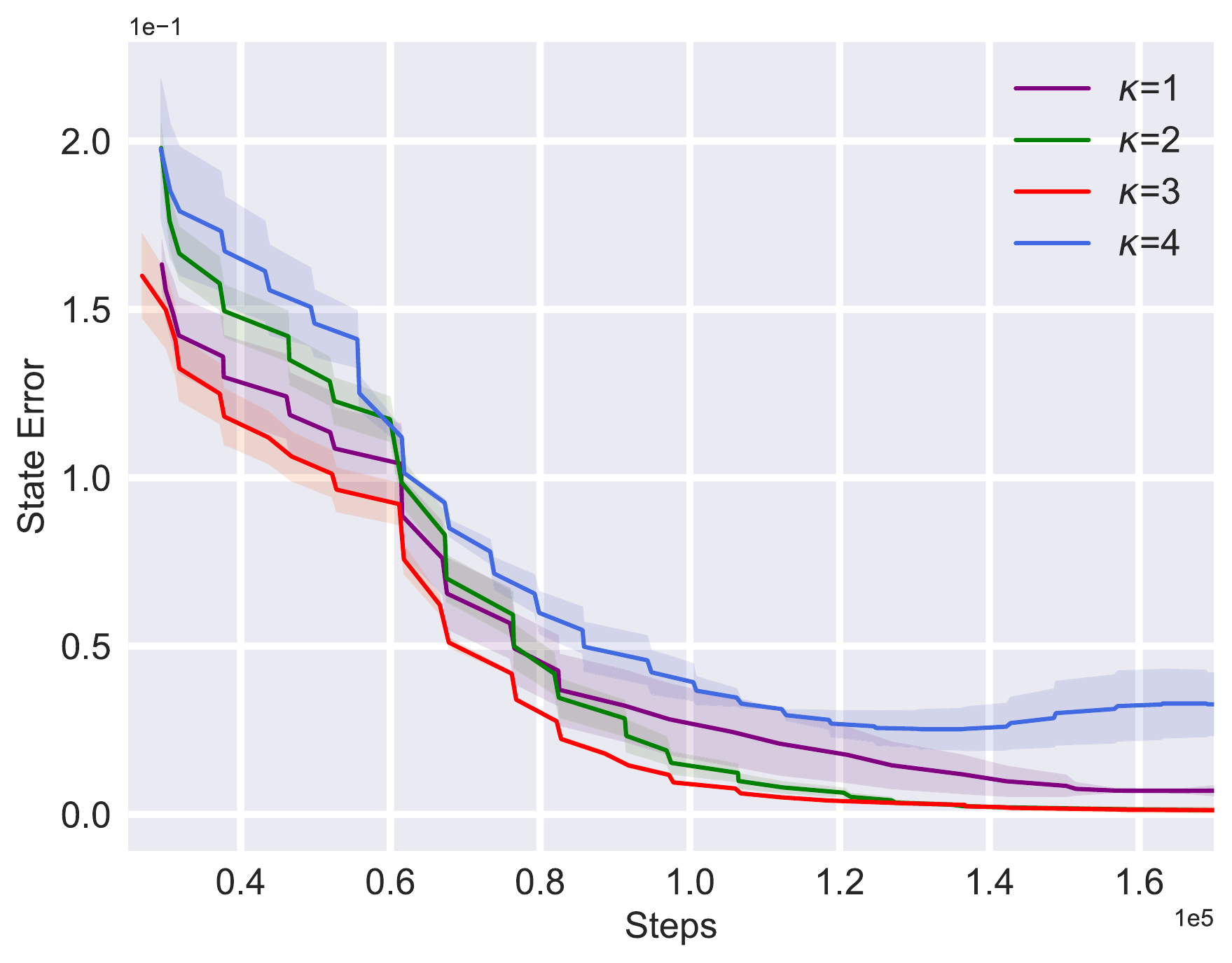}
    		}

    \caption{ Training performance on Ring Attenuation and CACC Catch-up under different choices of $\kappa$. (a) and (c) report training reward; (b) and (d) report the state error.
    } 
\label{fig:error}
\end{figure}

\begin{figure}[t!]
	\centering
	\subfigure[Episode reward in ATSC Grid]{
    		\includegraphics[width=0.46\columnwidth]{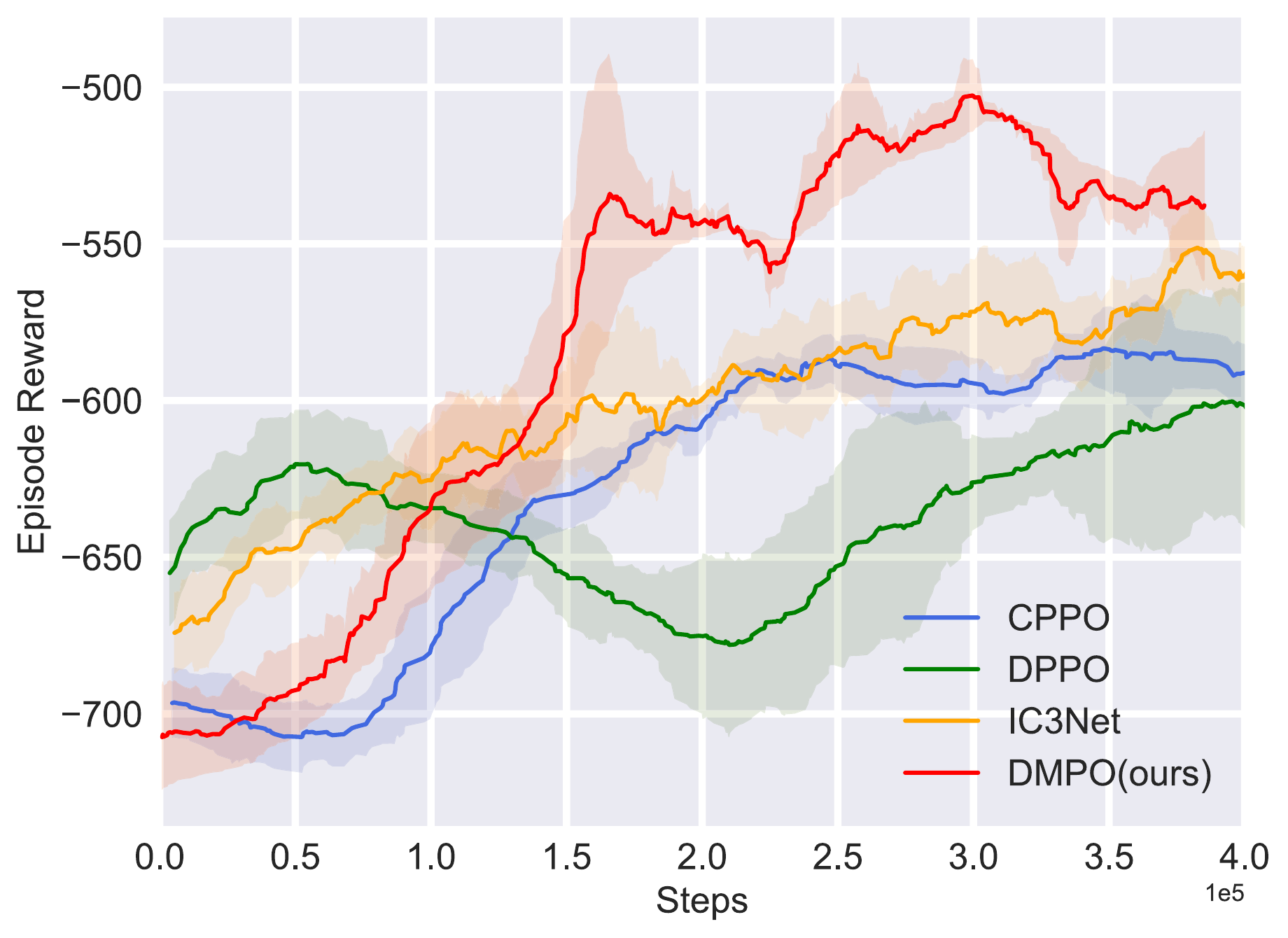}
    		}
	\subfigure[Model error in ATSC Grid]{
    		\includegraphics[width=0.44\columnwidth]{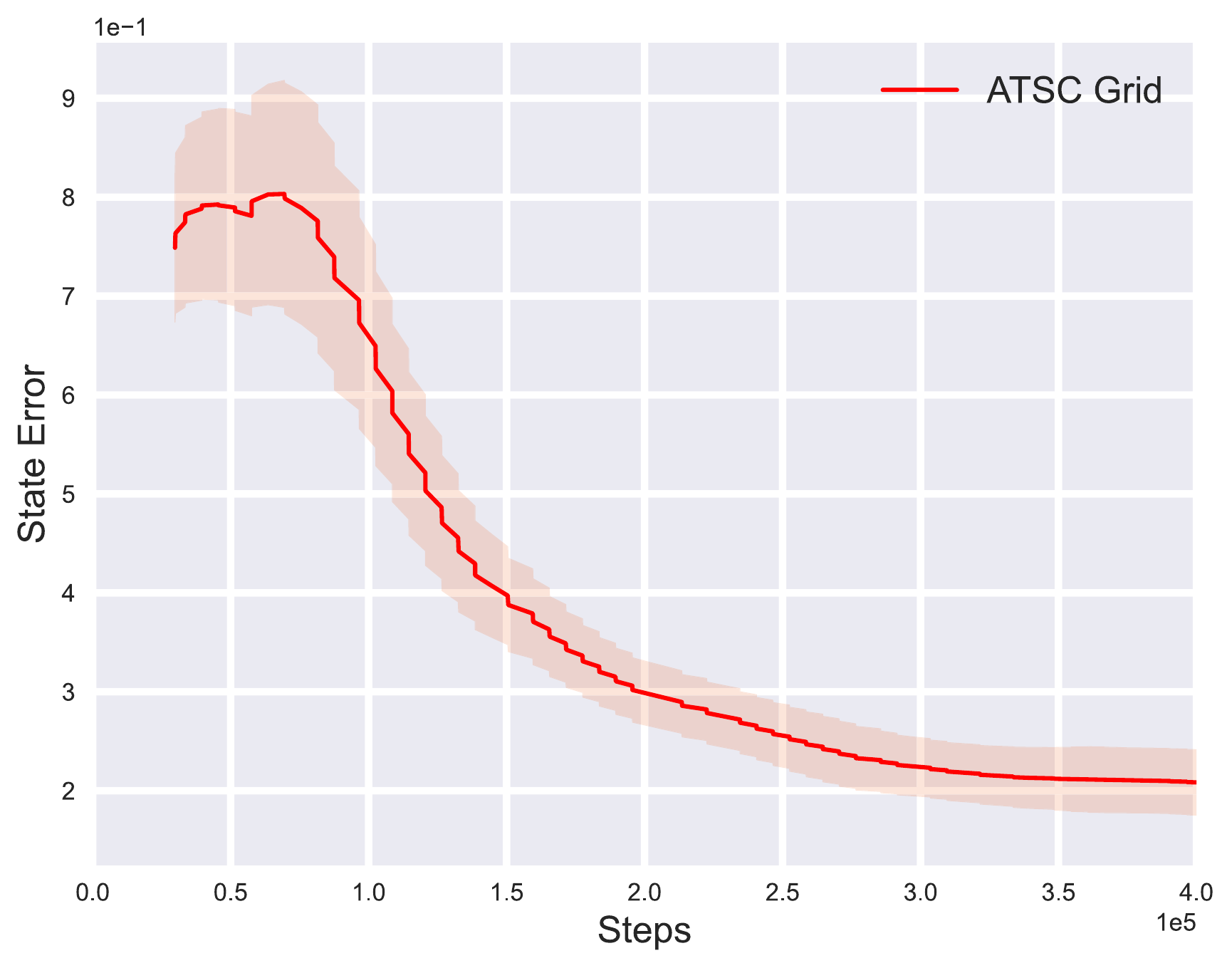}
    		}
    \caption{ Training curves on ATSC-Grid environment and state error in DMPO algorithm.
    } 
\label{fig:atsc}
\vspace{-0.2 in}
\end{figure}

{\textbf{Training Results}}
Figure \ref{fig:CACC} shows the results of episode reward v.s. number of training steps of different algorithms. 
From the results, we observe that our method achieves highest data efficiency, converging within $1e5$ training steps, exceedingly outperforms CPPO and DPPO.

The comparison between $\ourmethod$ and DPPO can be viewed as an ablation study of model usage. In figure eight, $\ourmethod$ increases sample efficiency at the beginning, but as the task becomes difficult, the sample efficiency of our method decreases. In a relatively easy task, ring attenuation, our method increased sample efficiency massively, compared with its model-free counterpart.

The comparison between the performance of CPPO and $\ourmethod$ or DPPO can be viewed as an ablation study of extended value function. From the result in four environments, we observe that the  performance of CPPO does not exceed that of the algorithms that use extended value function. In this way, we conclude that by using an extended value function, a centralized algorithm can be decomposed into a decentralized algorithm, but the performance would not drop significantly.

{Figure \ref{fig:error} shows training curves and the accuracy of our model in predicting the state during training under different choices of $\kappa$. The state error is defined as the MSE loss.
From the figures, we conclude that neighborhood information is accurate enough for a model to predict the next state in these environments, but the effect of different $\kappa$ on state prediction is different. When the network structure of DMPO is fixed, the sampling efficiency will increase and the error of our model in predicting the state will decrease with the increase of $\kappa$. But when $\kappa$ is greater than a certain value, the sampling efficiency will decrease and the error of our model in predicting the state will decrease due to the limited fitting ability of the network. The optimal choice of $\kappa$ is 3 under our network structure in both Ring Attenuation and CACC Catch-up.}

\subsection{Adaptive Traffic Signal Control}
{The objective of ATSC is to adaptively adjust signal phases to minimize traffic congestion based on real-time road-traffic measurements. Here we use the scenario: a 5x5 synthetic traffic grid. \textcolor{black}{The traffic grid is composed of two-lane roads.}
In the traffic grid, the peak hour traffic dynamics are simulated by a collection of four time-variant traffic flows, including loading and recovery phases. The state of each agent is all the twelve flows, and all agents have the same action space, which is a set of five pre-defined signal phases.}

{\textbf{Training results}}
{ Figure \ref{fig:atsc} shows training curves and the accuracy of our model in predicting the state during training in a relatively hard task with more agents, higher dimensional state space and more complex scenario settings. In ATSC, compared to the three model-free baselines, we conclude that the existence of the model increased sample efficiency massively. With training, the model error gradually decreases smoothly, indicating that the training process of the model is satisfactory and our model is more and more accurate in predicting future states.
}

\subsection{Execution Results}
\begin{figure}[t!]
	\centering
	\subfigure[Position profiles in Figure Eight]{
    		\includegraphics[width=0.45\columnwidth]{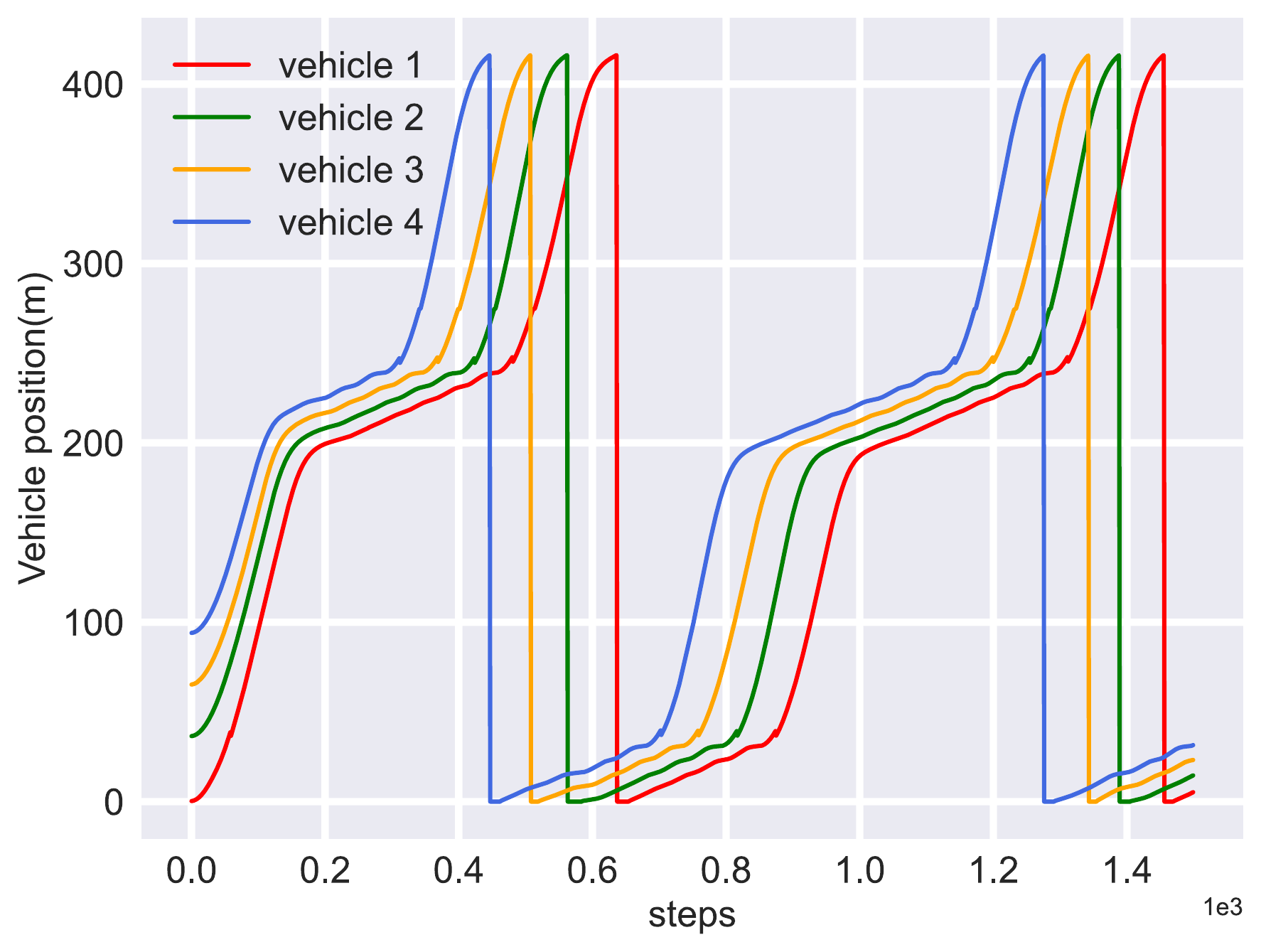}
    		}
	\subfigure[Velocity profiles in Figure Eight]{
    		\includegraphics[width=0.45\columnwidth]{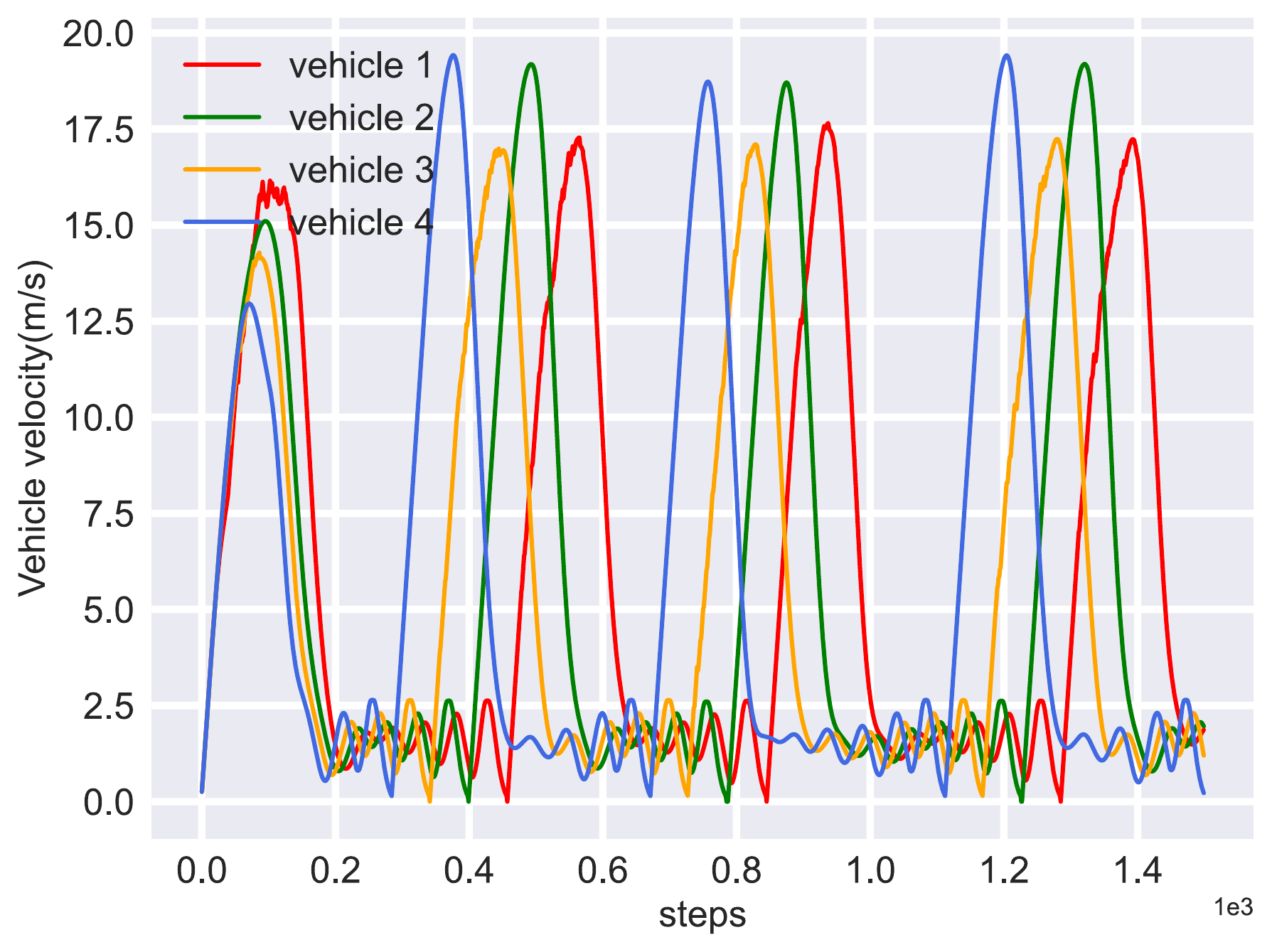}
    		}

    \subfigure[Headway profiles in CACC Catch-up]{
    		\includegraphics[width=0.44 \columnwidth]{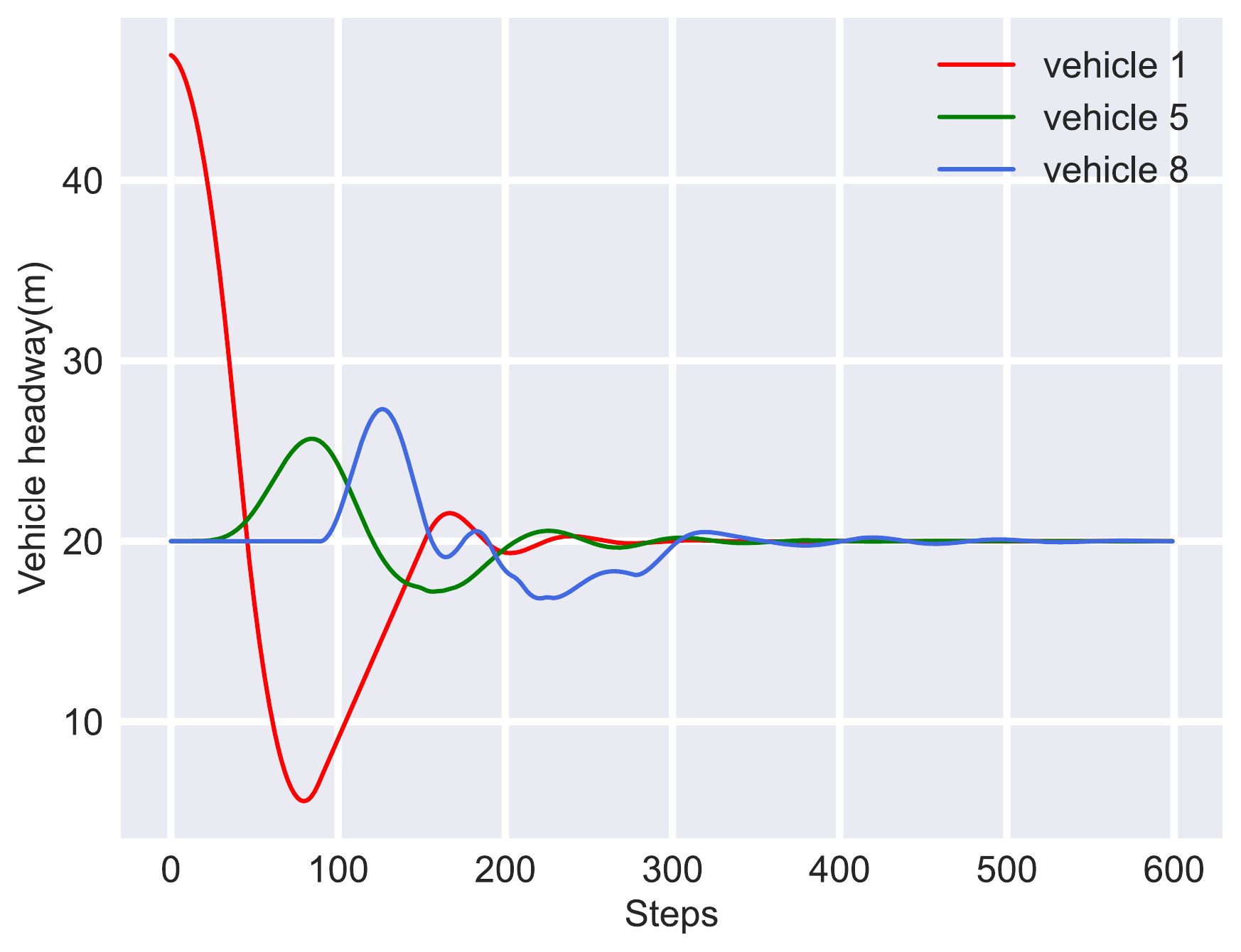}
    		}
    \subfigure[Velocity profiles in CACC Catch-up]{
    		\includegraphics[width=0.44 \columnwidth]{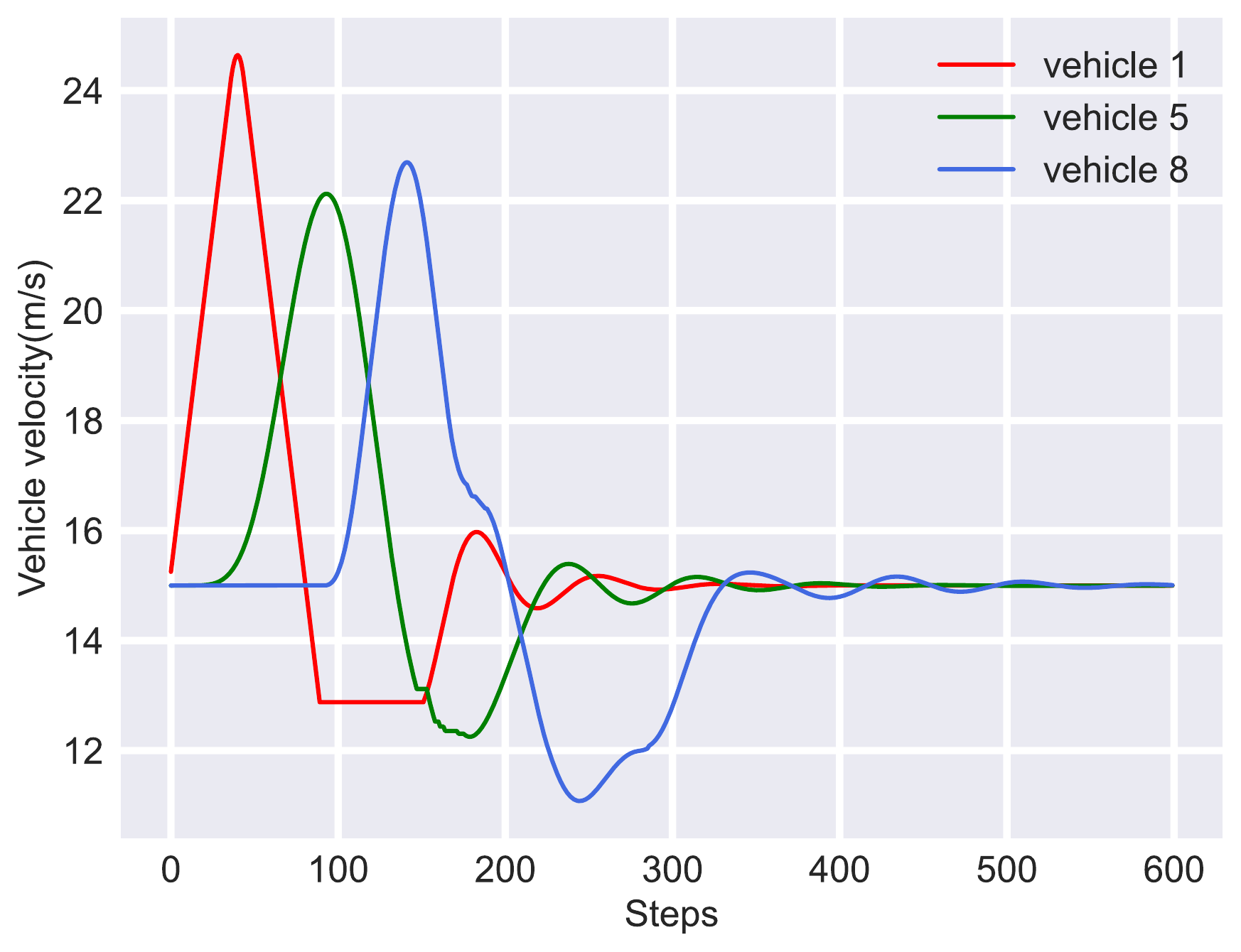}
    		}

    \caption{ Execution performance of the final decision models.
    } 
\label{fig:cacc-execution}
\vspace{-0.2 in}
\end{figure}

{
Figure \ref{fig:cacc-execution} shows execution performance of the trained policies based on DMPO. In (a) and (b), we show the position and velocity profiles of four adjacent vehicles 1, 2, 3, 4. 
Our models control the formation of queues to cross intersection with an orderly process of accelerating to the target velocity and then decelerating to the safe velocity near 0 $m/s$. We conclude that our models control vehicles to obey traffic rules while improving the efficiency of the overall traffic flow.}

{
In CACC Catch-up, we plot the headway and velocity profiles of three vehicles 1,5,8 at the head, middle and tail. 

The  headway stabilizes at 20 $m$ which matches the target headway. 
 The velocity stabilizes at  15 $m/s$, matching the target velocity.
Both the velocity and headway of three vehicles with the same interval in queue can be stably controlled around the target.
}


%
\section{Conclusions} \label{sec:conclusion}
In this paper, we propose $\ourmethod$, a model-based and decentralized multi-agent RL framework. 
To preserve the independence of policy learning, we introduce extended value function and the resulting policy gradient is proven to be a close approximation to true policy gradient.
Through extensive experiments in several tasks in networked systems, we show that our algorithm matches the performance of some state-of-art multi-agent algorithms and achieves higher data efficiency. From the results, we also conclude that using extended value function instead of centralized value function did not sacrifice performance massively, yet it makes our algorithm scalable. 
One limitation of this work is that  the system is assumed to be factorizable into independent components, which may not hold in practice. We leave the more general scenarios for future work. 




\section*{APPENDIX} \label{app:exp_detail}

\subsection{Proof of Theorem \ref{thm:V}}

\begin{proof}
In \cite{qu2020scalable2}, it was proven that if $s_{N_i^\kappa}$ and $a_{N_i^\kappa}$ are fixed, then no matter how distant states and actions changes, $Q$-function will not change significantly:
\begin{equation*} \label{eq:trunc_Q}
\resizebox{\columnwidth}{!}{
    $|Q_i(s_{N_i^\kappa},a_{N_i^\kappa},s_{N_{-i}^\kappa},a_{N_{-i}^\kappa})-Q_i(s_{N_i^\kappa},a_{N_i^\kappa},s_{N_{-i}^\kappa}',a_{N_{-i}^\kappa}')|\leq\frac{r_{\max}}{1-\gamma}\gamma^\kappa.$
    }
\end{equation*}

As value function is the expectation of Q-function
\begin{equation*}
\begin{aligned}
    & V_i(s) = \mathbb{E}_{a\sim\pi}Q_i(s,a) \\
    & V_i(s_{N_i^\kappa}) = \mathbb{E}_{a\sim\pi}Q_i(s_{N_i^\kappa},a_{N_i^\kappa}),
\end{aligned}
\end{equation*}
we have,
\begin{equation*}
\begin{aligned}
    |V_i(s)-V_i(s_{N_i^\kappa})| & = |\mathbb{E}_{a\sim\pi}Q_i(s,a) - \mathbb{E}_{a\sim\pi}Q_i(s_{N_i^\kappa},a_{N_i^\kappa})| \\
    & \leq \mathbb{E}_{a\sim\pi}|Q_i(s,a) - Q_i(s_{N_i^\kappa},a_{N_i^\kappa})| \\
    & \leq \frac{r_{\max}}{1-\gamma}\gamma^\kappa,
\end{aligned}
\end{equation*}
which concludes the proof.
\end{proof}

\subsection{Proof of Theorem \ref{thm:pg}}
\begin{proof}
The difference of the gradients is written as
\begin{equation} \label{eq:grad1}\small
\begin{aligned}
    & g_i - \tilde{g_i} =  \mathbb{E}(\hat{A}-\tilde{A})\nabla_{\theta_i}\log\pi_i(a_i|s_{ {N}_i}) \\
    = & \frac{1}{n}\mathbb{E}[\sum_{j\notin N_i^\kappa}\hat{A}_j] \nabla_{\theta_i}\log\pi_i(a_i|s_{ {N}_i}) 
    \\
    &+ \frac{1}{n}\mathbb{E}[\sum_{j\in N_i^\kappa}(\hat{A}_j-\tilde{A}_j)] \nabla_{\theta_i}\log\pi_i(a_i|s_{ {N}_i}) \\
    = & \frac{1}{n}\mathbb{E}[\sum_{j\notin N_i^\kappa}(r_j+\gamma V_j(s')-V_j(s))] \nabla_{\theta_i}\log\pi_i(a_i|s_{{N}_i}) \\
      & + \frac{1}{n}\mathbb{E}\sum_{j\in N_i^\kappa}\big[(r_j+\gamma V_j(s')-V_j(s))\\
      &\quad \  -(r_j+\gamma V_j(s_{N_i^\kappa}')-V_j(s_{N_i^\kappa}))\big] \nabla_{\theta_i}\log\pi_i(a_i|s_{ {N}_i}) \\
    = & L_1 + L_2.
\end{aligned}
\end{equation}
Note that for any function $b(s)$,
$\mathbb{E}[b(s)\nabla_{\theta_i}\log\pi^{\theta_i}(a_i|s_{{N}_i})]=0$. Therefore, $L_2$ in Equation \eqref{eq:grad1} becomes:
\begin{equation} \label{eq:l1}
\begin{aligned}
    |L_2| \leq & \frac{1}{n}\mathbb{E}\sum_{j\in N_i^\kappa}\gamma|V_j(s')-V_j(s_{N_i^\kappa}')]||\nabla_{\theta_i}\log\pi_i(a_i|s_{{N}_i})|| \\
    \leq & \frac{|N_i^\kappa|}{n}\frac{\gamma^{\kappa+1}}{1-\gamma}r_{\max}g_{\max}.
\end{aligned}
\end{equation}
For $L_1$, note that {$r_j+\gamma V_j(s')-V_j(s)=-V_j(s)+r_j+\mathbb{E}\sum_{t=1}^{\kappa-2}\gamma^t r_j^t + \gamma^{\kappa-1}V_j(s^{\kappa-1})$}. And in an independent network system (where Eq.\eqref{eq:factorization-nsc} holds), $s_j^t,a_j^t,t=1,...,\kappa-2$ is not affected by policy $\pi_i$ if $j\notin N_i^\kappa$, we have that
\begin{equation} \label{eq:l2}
\begin{aligned}\small
    |L_1| \leq & \frac{1}{n}\mathbb{E}\sum_{j\notin N_i^\kappa}|\gamma^{\kappa-1}V_j(s^{\kappa-1})| |\nabla_{\theta_i}\log\pi_i(a_i|s_{{N}_i})|| \\
    \leq & (1-\frac{|N_i^\kappa|}{n})\frac{\gamma^{\kappa-1}}{1-\gamma}r_{\max}g_{\max}.
\end{aligned}
\end{equation}
Put Equation \eqref{eq:l1} and \eqref{eq:l2} together, we have
\begin{equation*}
\begin{aligned}
    |g_i-\tilde{g}_i|\leq & |L_1| + |L_2| \\
    \leq & \frac{|N_i^\kappa|}{n}\frac{\gamma^{\kappa+1}}{1-\gamma}r_{\max}g_{\max} + (1-\frac{|N_i^\kappa|}{n})\frac{\gamma^{\kappa-1}}{1-\gamma}r_{\max}g_{\max} \\
    = & \frac{\gamma^{\kappa-1}}{1-\gamma}[1-(1-\gamma^2)\frac{|N_i^\kappa|}{n}]r_{\max}g_{\max}.
\end{aligned}
\end{equation*}
\end{proof}



\balance
\bibliographystyle{IEEEtran}
\bibliography{main}

\end{document}